\documentclass[10pt,twocolumn,letterpaper]{article}

\usepackage[pagenumbers]{cvpr} % To force page numbers, e.g. for an arXiv version

\usepackage{graphicx}
\usepackage{amsmath}
\usepackage{amssymb}
\usepackage{booktabs}

\usepackage{comment}
\usepackage{color}
\usepackage{epsfig}
\usepackage{tabularx}
\usepackage{booktabs}
\usepackage{multirow}
\usepackage{bm}
\usepackage{algorithm,algpseudocode}
\usepackage{xcolor,colortbl}
\usepackage[pagebackref,breaklinks,colorlinks]{hyperref}
\usepackage{breakcites}
\usepackage{dsfont}
\usepackage{bbm}
\usepackage{makecell}
\usepackage{colortbl}
\usepackage{marvosym}
\usepackage[accsupp]{axessibility}

\usepackage[capitalize]{cleveref}
\crefname{section}{Sec.}{Secs.}
\Crefname{section}{Section}{Sections}
\Crefname{table}{Table}{Tables}
\crefname{table}{Tab.}{Tabs.}

\def\cvprPaperID{1351} % *** Enter the CVPR Paper ID here
\def\confName{CVPR}
\def\confYear{2023}

\definecolor{myblue}{RGB}{23,183,241}
\definecolor{mygray}{RGB}{230,230,230}
\definecolor{myblue2}{RGB}{198,213,250}
\definecolor{mygreen}{RGB}{197,224,180}
\definecolor{myyellow}{RGB}{255,230,153}
\definecolor{myorange}{RGB}{244,171,131}

\begin{document}


% \makeatletter
% \renewcommand*{\@fnsymbol}[1]{\ensuremath{\ifcase#1\or *\or \dagger \or \ddagger\or
%   \mathsection\or \mathparagraph\or \|\or **\or \dagger\dagger
%   \or \ddagger\ddagger \else\@ctrerr\fi}}
% \makeatother

\title{PLA: Language-Driven Open-Vocabulary 3D Scene Understanding}

\author{Runyu Ding$^1$\footnotemark[1]~\footnotemark[2]~~ Jihan Yang$^{1*}$~~ Chuhui Xue$^2$~~ Wenqing Zhang$^2$~~ Song Bai$^{2}$\footnotemark[3]~~ Xiaojuan Qi$^{1}$\footnotemark[3]\\
$^1$The University of Hong Kong~~~ $^2$ByteDance\\
% {\tt\small \{ryding,jhyang,xjqi\}@eee.hku.hk ~~ xuec0003@e.ntu.edu.sg} \\ {\tt\small wenqingzhang@bytedance.com ~~ songbai.site@gmail.com}
}

\maketitle
\renewcommand{\thefootnote}{\fnsymbol{footnote}}
\footnotetext[1]{Equal contribution: \{ryding, jhyang\}@eee.hku.hk}
% $^\ddagger$Corresponding authors.}
\footnotetext[2]{Part of the work is done during an internship at ByteDance AI Lab.}
\footnotetext[3]{Corresponding authors: song.site@gmail.com, xjqi@eee.hku.hk}

%%%%%%%%% ABSTRACT
\begin{abstract}
    Open-vocabulary scene understanding aims to localize and recognize unseen categories beyond the annotated label space.
    The recent breakthrough of 2D open-vocabulary perception is largely driven by Internet-scale paired image-text data with rich vocabulary concepts. However, this success cannot be directly transferred to 3D scenarios due to the inaccessibility of large-scale 3D-text pairs. To this end, we propose to distill knowledge encoded in pre-trained vision-language (VL) foundation models through captioning multi-view images from 3D, which allows explicitly associating 3D and semantic-rich captions. Further, to foster coarse-to-fine visual-semantic representation learning from captions, we design hierarchical 3D-caption pairs, leveraging geometric constraints between 3D scenes and multi-view images. Finally, by employing contrastive learning, the model learns language-aware embeddings that connect 3D and text for open-vocabulary tasks. 
    Our method not only remarkably outperforms baseline methods by 25.8\% $\sim$ 44.7\% hIoU and 14.5\% $\sim$ 50.4\% hAP$_{50}$ in open-vocabulary semantic and instance segmentation, but also shows robust transferability on challenging zero-shot domain transfer tasks. See the project website at  \href{https://dingry.github.io/projects/PLA}{ https://dingry.github.io/projects/PLA}.
\end{abstract} 

%%%%%%%%% BODY TEXT
% \vspace{-0.4cm}
\section{Introduction}
\vspace{-0.2cm}
\label{sec:intro}

\begin{figure}[t]
    % \vspace{-0.3cm}
    \begin{center}
    \includegraphics[width=1\linewidth]{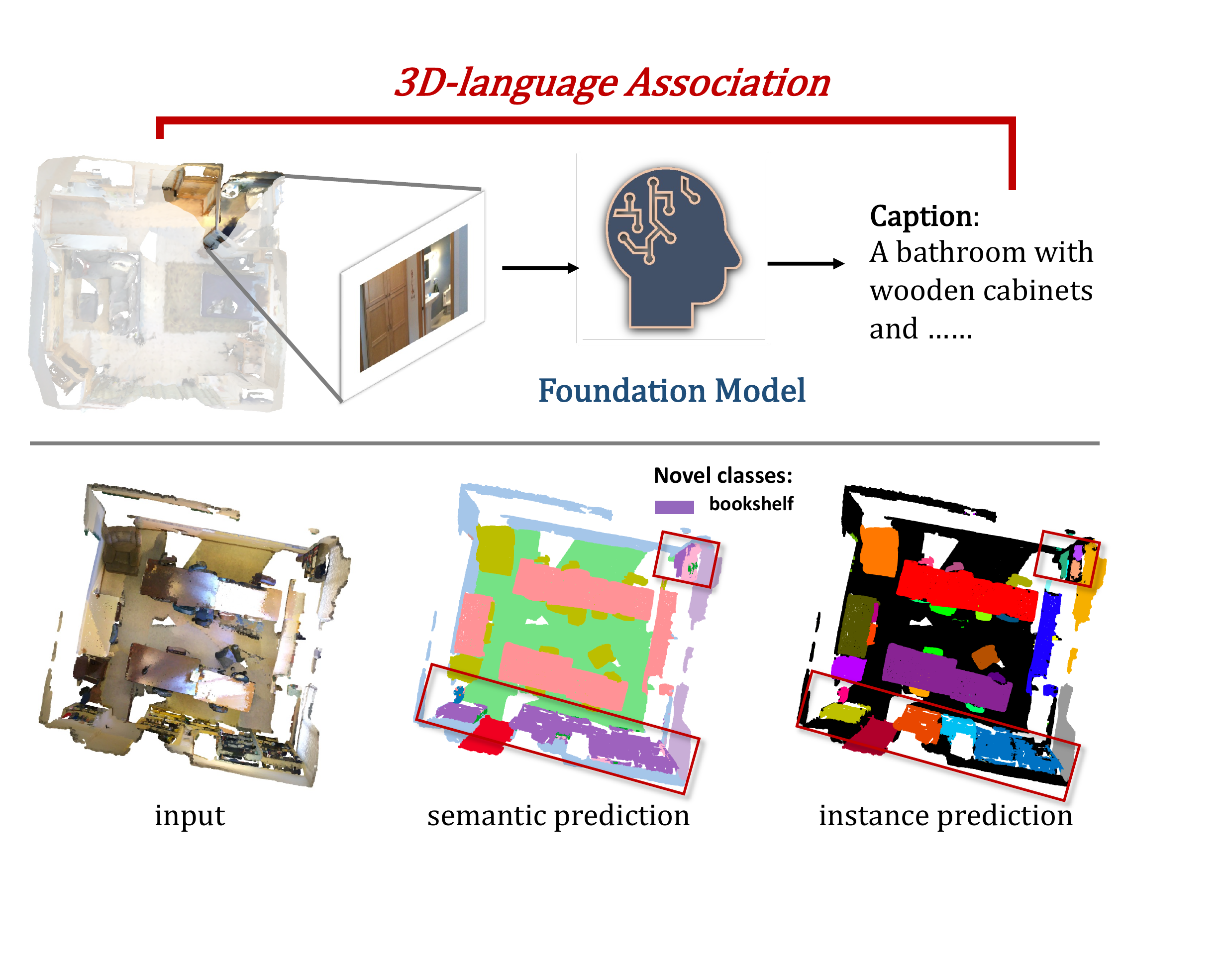}
    \end{center}
    \vspace{-0.6cm}
    % captionsetup{font={small}}
    \caption{An example of 3D open-vocabulary scene understanding with ``bookshelf'' as unseen class for ScanNet~\cite{dai2017scannet}. The close-set model mistakes ``bookshelf'' as ``cabinet'' or simply misses ``bookshelf'' in (a) and (c). Our open-vocabulary model correctly localizes and recognizes ``bookshelf'' in (b) and (d).}
    \vspace{-0.6cm}
    \label{fig:teaser}
\end{figure}

3D scene understanding is a fundamental perception component in real-world applications such as robot manipulation, virtual reality and human-machine interaction. 
Deep learning has attained remarkable success in this area~\cite{graham20183d,vu2022softgroup,misra2021-3detr}. However, deep models trained on a human-annotated dataset are only capable of understanding semantic categories in that dataset, \ie closet-set prediction. As a result, they fail to recognize unseen categories in the open world (see Fig.~\ref{fig:teaser}). 
This largely restricts their applicability in real-world scenarios with unbounded categories.
Besides, heavy annotation costs on 3D datasets (\eg 22.3 minutes for one scene with 20 classes~\cite{dai2017scannet}) further make it infeasible to rely on  human labor to cover all real-world categories.

This motivates us to study open-vocabulary 3D scene understanding, which equips a  model with the ability to localize and recognize open-set classes beyond the label space of an annotated dataset (see Fig.~\ref{fig:teaser}).
Recently, vision-language (VL) foundation models ~\cite{radford2021learning,jia2021scaling,florence} trained on billions of web-crawled image data with semantic-rich captions~\cite{sharma2018conceptual} are capable of learning adequate vision-language embeddings to connect text and image, which are further leveraged to solve many 2D open-vocabulary tasks including object detection~\cite{gu2021open,Hanoona2022Bridging}, semantic segmentation~\cite{xu2021simple,li2022languagedriven,zhou2022maskclip}, visual question answering~\cite{piergiovanni2022answer} and \etc. 
Albeit significantly advancing open-vocabulary image understanding tasks, this pre-training paradigm is not directly viable in the 3D domain due to the absence of large-scale 3D-text pairs. 

To this end, initial efforts~\cite{zhang2022pointclip,huang2022clip2point} have attempted to project 3D data into 2D modalities, such as RGB images and depth maps, enabling pre-trained VL foundation models to process the 2D data and achieve object-level open-vocabulary recognition. Nevertheless, this line of methods suffers from several major issues, making it suboptimal to handle scene-level understanding tasks (e.g., instance segmentation). 
First, multiple RGB images and depth maps are required to represent a 3D sample, which incurs heavy computation and memory costs during training and inference. Second, the projection from 3D to 2D induces information loss and prohibits direct learning from rich 3D data, leading to subpar performance. Our preliminary study shows the cutting-edge 2D open-vocabulary semantic segmentation method MaskCLIP \cite{zhou2022maskclip} attains a mere $17.8\%$ mIoU with a 20-fold increase in latency when applied to analyze projected 2D images from 3D ScanNet dataset.

Thus, considering the success of VL foundation models for a variety of vision-language tasks~\cite{gu2021open,Hanoona2022Bridging,xu2021simple,li2022languagedriven,zhou2022maskclip,zhang2022pointclip,huang2022clip2point}, we ask: \textit{is it possible to elicit knowledge encoded in powerful VL foundation models to build an explicit association between 3D and language for open-vocabulary understanding?}  
To this end, our core idea is to exploit pre-trained VL foundation models~\cite{vit-gpt2,wang2022ofa} to caption easily-obtained image data aligned with 3D data ({\ie} the point set in the corresponding frustum to produce the image). Note that these images can be acquired through neural rendering~\cite{dai2020neural,yu2021pixelnerf} or from the 3D data collection pipeline~\cite{dai2017scannet}. By doing so, we can distill semantic-rich textual descriptions to the 3D domain, which allows explicit association between 3D and vocabulary-rich text for zero-shot 3D scene understanding. 

Given 3D-language association, the next question is enabling a 3D network to learn language-aware embeddings from (pseudo) captions. The key challenge stems from intricate object compositions in 3D scene-level data (see Fig.~\ref{fig:caption}), making it difficult to connect objects with corresponding words in the caption.  
This differs from object-centric image data containing a single centered object~\cite{radford2021learning}. 
Fortunately, the captioned multi-view images from a 3D scene are related by 3D geometry, which can be leveraged to build hierarchical point-caption pairs, including scene-, view- and entity-level captions. These multi-level point-caption pairs offer coarse-to-fine supervision signals, facilitating learning adequate visual-semantic representations from rich vocabulary by contrastive learning. 
% We name our method PLA, \ie point-language association \ry{still thinking ...}.
Without task-specific design, our \textbf{P}oint-\textbf{L}anguage \textbf{A}ssociation paradigm, namely PLA, is generic for various open-vocabulary 3D scene understanding tasks, such as semantic and instance segmentation.
% Without task-specific design, our language-driven paradigm is generic for various open-vocabulary 3D scene understanding tasks, \eg semantic and instance segmentation.

Experimental results for ScanNet~\cite{dai2017scannet} and S3IDS~\cite{armeni20163d} datasets show the effectiveness of our method in in-domain open-vocabulary tasks with only category shifts, {\ie} training and evaluation are conducted on the same dataset, surpassing baselines by 25.8\% $\sim$ 44.7\% hIoU on semantic segmentation and 14.5\% $\sim$ 50.4\% hAP$_{50}$ on instance segmentation. 
Besides, our model, trained on a dataset ({\ie} ScanNet), can generalize to another dataset ({\ie} S3IDS) with both data distribution and category shifts, manifesting its transferability. 
Finally, our model can benefit from more advanced foundation models that provide higher-quality caption supervision, showing its scalability and extensibility.

%-------------------------------------------------------------------------
\vspace{-0.2cm}
\section{Related Work}
\vspace{-0.2cm}
\noindent\textbf{3D scene understanding} focuses on understanding the semantic meaning of objects and surrounding environment from point clouds. 
In this work, we  focus on two fundamental scene understanding tasks: semantic and instance segmentation.
\textit{3D semantic segmentation} aims to obtain point-wise semantic predictions for point clouds. Representative works develop point-based solutions \cite{qi2017pointnet++,huang2018recurrent} with elaborately designed point convolution operations~\cite{thomas2019KPConv,xu2021paconv} or transformers~\cite{lai2022stratified}  or voxel-based ~\cite{graham20183d,choy20194d} methods with 3D sparse convolutions \cite{graham2017submanifold} to produce point-wise segmentation results. 
\textit{3D instance segmentation} further targets distinguishing different object instances based on semantic segmentation. 
Existing approaches either adopt a top-down solution~\cite{yi2019gspn,yang2019learning} via predicting 3D bounding box followed by mask refinement, or a  bottom-up~\cite{jiang2020pointgroup,vu2022softgroup} approach through grouping points.
However, existing methods cannot recognize open-set novel categories, which we aim to address.

\vspace{0.05in}\noindent\textbf{Zero-shot and open-vocabulary understanding} aims to recognize novel classes that are not annotated in training data. Early approaches mainly follow zero-shot settings that can be coarsely grouped into discriminative methods~\cite{xian2019semantic,baek2021exploiting} and generative  methods~\cite{bucher2019zero,gu2020context}. 3DGenZ~\cite{michele2021generative} extends~\cite{bucher2019zero} to the 3D scenario for zero-shot semantic segmentation.  
Going beyond zero-shot learning,
the more general open-vocabulary setting assumes a large vocabulary corpus is accessible during training~\cite{zareian2021open}.
Existing \textit{2D open-vocabulary learning} works either exploit massive annotated image-text pairs to provide weak supervision for expanding vocabulary size~\cite{zareian2021open,zhou2022detecting} or leverage pre-trained VL models from large-scale image-caption pairs, such as CLIP~\cite{radford2021learning}, to address open-vocabulary recognition where knowledge distillation~\cite{Hanoona2022Bridging,gu2021open,zang2022open} and prompt learning~\cite{feng2022promptdet,du2022learning} are studied.

In comparison, \textit{3D open-vocabulary learning} is still in its infancy with only a few explorations focusing on object classification~\cite{zhang2022pointclip,huang2022clip2point}. 
They attempt to project object-level 3D point clouds to multi-view 2D images and depth maps to adopt the pre-trained VL model to generate open-vocabulary predictions, which, however, suffer from heavy computation and poor performance if applied to 3D scene understanding tasks. 
In this work, we propose a language-driven 3D open-vocabulary framework that directly associates 3D with text descriptions leveraging multi-view images and VL foundation models. It can be generally applied to various scene understanding tasks and is efficient with only the 3D network employed in training and inference.

%-------------------------------------------------------------------------
\vspace{-0.1cm}
\section{Method}
\vspace{-0.1cm}

\begin{figure*}[t]
    % \vspace{-0.3cm}
    \begin{center}
    \includegraphics[width=1\linewidth]{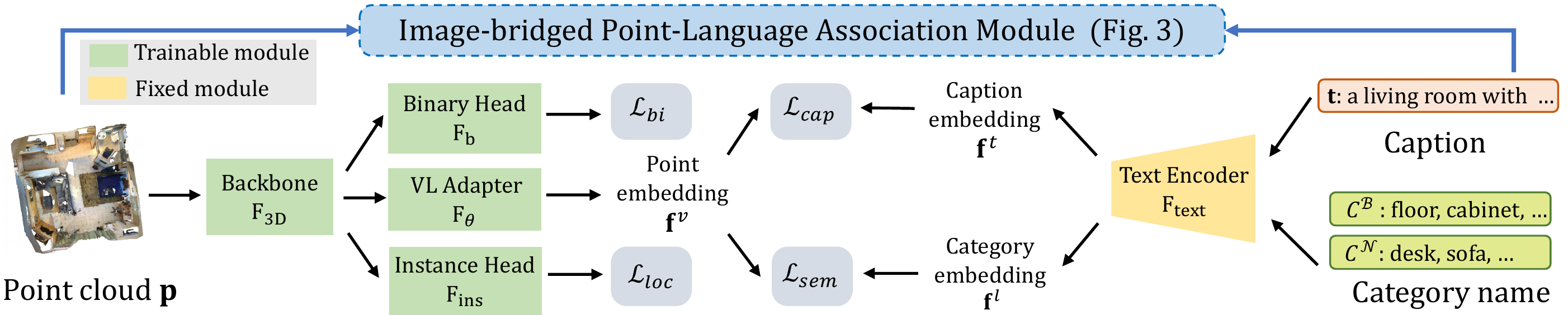}
    \end{center}
    \vspace{-0.5cm}
    % captionsetup{font={small}}
    \caption{Our language-driven 3D scene understanding paradigm. Different from the close-set network, the learnable semantic head is replaced by category embeddings encoded by a text encoder from category names. Binary head is to rectify semantic scores with base and novel probability as conditions. Instance head is tailored to instance segmentation. Most importantly, to endow the model with rich semantic space to improve open-vocabulary capability, we supervise point embeddings with caption embeddings based on point-language association (see Fig.~\ref{fig:caption} for details). Best viewed in color.}
    \vspace{-0.4cm}
    \label{fig:framework}
    \end{figure*}

\subsection{Preliminary}\label{sec:method_pre}
\vspace{-0.1cm}
3D open-vocabulary scene understanding aims to localize and recognize unseen categories without corresponding human annotation as supervision. 
Formally, annotations on semantic and instance levels $\mathcal{Y} = \{\mathbf{y}^{\text{sem}}, \mathbf{y}^{\text{ins}}\}$ are divided into base $\mathcal{C}^B$ and novel $\mathcal{C}^N$ categories. In the training stage, the 3D model can access all point clouds $\mathcal{P}=\{\textbf{p}\}$ but only annotations for base classes $\mathcal{Y}^B$, unaware of both annotations $\mathcal{Y}^N$ and category names concerning novel classes $\mathcal{C}^N$.
However, during inference, the 3D model needs to localize objects and classify points belonging to both base and novel $\mathcal{C}^B \cup \mathcal{C}^N$ categories.

As for a typical scene understanding network, it consists of a 3D encoder $\text{F}_{\text{3D}}$, a dense semantic classification head $\text{F}_{\text{sem}}$ and an instance localization head $\text{F}_{\text{loc}}$ (see Suppl. for details). Its inference pipeline can be demonstrated below,
\vspace{-0.2cm}
\begin{equation}
    \vspace{-0.2cm}
    \mathbf{f}^p=\text{F}_{\text{3D}}(\mathbf{p}), ~~
    \mathbf{s}= \sigma \circ \text{F}_{\text{sem}}(\mathbf{f}^p), ~~
    \mathbf{z}=\text{F}_{\text{loc}}(\mathbf{f}^p,  \mathbf{s}),
\end{equation}
where $\mathbf{p}$ is the input point cloud, $\mathbf{f}^p$ is point-wise visual feature, $\mathbf{s}$ is semantic score, $\mathbf{z}$ is the instance proposal output and $\sigma$ is the softmax function. With these network predictions, we can then calculate semantic classification loss $\mathcal{L}_\text{sem}$ with semantic label $\mathbf{y}^\text{sem}$, and localization loss $\mathcal{L}_\text{loc}$ with instance label $\mathbf{y}^\text{ins}$  similar to~\cite{jiang2020pointgroup,vu2022softgroup} as Eq~\eqref{eq:basic_loss}. Notice that $\mathbf{y}^\text{sem}$ and $\mathbf{y}^\text{ins}$ only relate to base categories $\mathcal{C}^B$. 
\vspace{-0.2cm}
\begin{gather}\label{eq:basic_loss}
\vspace{-0.3cm}
    \mathcal{L}_{\text{sem}}=\text{Loss}(\mathbf{s}, \mathbf{y}^\text{sem}), ~~ \mathcal{L}_{\text{loc}}=\text{Loss}(\mathbf{z}, \mathbf{y}^\text{ins}).
    % \vspace{-0.2cm}
\end{gather}

\subsection{Open-Vocabulary Setups}
\vspace{-0.1cm}
Though we can train a scene understanding model with loss functions in  Eq.~\eqref{eq:basic_loss}, it is actually a close-set model with a close-set classifier $\text{F}_{\text{sem}}$, incapable of recognizing unseen categories.
%in the inference stage.
In this regard, we introduce the text-embedded semantic classifier to obtain an open-vocabulary model and propose a binary calibration module to correct 
the bias toward base categories for open-vocabulary inference.
\vspace{-0.3cm}
\subsubsection{Text-Embedded Semantic Classifier}\label{sec:text_embed_cls}
\vspace{-0.1cm}
First, as shown in Fig.~\ref{fig:framework}, to make the model become an open-vocabulary learner, we %follow previous 2D paradigms~\cite{Hanoona2022Bridging,gu2021open} to 
replace its learnable semantic classifier $\text{F}_\text{sem}$ with category embeddings $\mathbf{f}^l$ and a learnable vision-language adapter $\text{F}_\theta$ to match the dimension between 3D features $\mathbf{f}^p$ and $\mathbf{f}^l$ as follows, 
\vspace{-0.2cm}
\begin{equation}\label{eq:text_encoder}
    \vspace{-0.2cm}
    \mathbf{f}^v=\text{F}_\theta(\mathbf{f}^p), ~~ \mathbf{s}= \sigma(\mathbf{f}^l\cdot \mathbf{f}^v),
\end{equation}
where $\mathbf{f}^v$ is the projected features with the VL adapter $\text{F}_\theta$, $\mathbf{f}^l = [\mathbf{f}^l_1, \mathbf{f}^l_2, \cdots, \mathbf{f}^l_{k}]$ is a series of category embeddings obtained by encoding category names $\mathcal{C}$ with a frozen text encoder $\text{F}_{\text{text}}$ such as BERT~\cite{devlin2018bert} or CLIP~\cite{radford2021learning} (see Fig.~\ref{fig:framework}). The prediction is made by calculating the 
cosine similarity among projected point features $\mathbf{f}^v$ and 
categories $\mathbf{f}^l$ and then selecting the most similar category. 
Notice that $\mathbf{f}^l$ only contains embeddings belonging to base categories $\mathcal{C}^B$ during training, but embeddings related to both base and novel classes $\mathcal{C}^B \cup \mathcal{C}^N$ are used during open-vocabulary inference.
With category embeddings $\mathbf{f}^l$ as a classifier, the model can support open-vocabulary inference with any desired categories. The above design generally follows LSeg~\cite{li2022languagedriven} and is named LSeg-3D as a baseline.

\vspace{-0.3cm}
\subsubsection{Semantic Calibration with Binary Head}
\vspace{-0.1cm}
Although the model has already possessed open-vocabulary capability, we empirically find that it can hardly make any correct predictions on novel classes but mistakes them for base classes. 
As the model is only trained to recognize base categories, it inevitably produces over-confident predictions on base classes regardless of their correctness, also known as the calibration problem~\cite{guo2017calibration}.
To this end, we propose a binary calibration module to rectify semantic scores with the probability of a point belonging to base or novel classes. 

Specifically, as shown in Fig.~\ref{fig:framework}, the binary head $\text{F}_\text{b}$ is employed to distinguish annotated ({\ie} base) and unannotated ({\ie} novel) points. During training, $\text{F}_\text{b}$ is optimized with:
\vspace{-0.2cm}
\begin{equation}
    \vspace{-0.2cm}
    \mathbf{s}^b = \text{F}_\text{b}(\mathbf{f}^p), ~~ \mathcal{L}_{bi}=\text{BCELoss}(\mathbf{s}^b, \mathbf{y}^b),
\end{equation}
where BCELoss($\cdot$, $\cdot$) is the binary cross-entropy loss, $\mathbf{y}^b$ is the binary label and $\mathbf{s}^b$ is the predicted binary score indicating the probability that a point belongs to novel categories. 
In the inference stage, we then exploit the binary probability 
$\mathbf{s}^b$ to correct the over-confident semantic score $\mathbf{s}$ as follows,
\vspace{-0.2cm}
\begin{equation}\label{eq:binary}
\vspace{-0.2cm}
    \mathbf{s} = \mathbf{s}_B\cdot (1-\mathbf{s}^b)+\mathbf{s}_N\cdot \mathbf{s}^b,
\end{equation}
where $\mathbf{s}_B$ is the semantic score computed solely on base classes with novel class scores set to zero.  
Similarly, $\mathbf{s}_N$ is computed only for novel classes, setting base class scores to zero.
We empirically show that the probability calibration largely improves the performance of both base and novel categories (see Sec.~\ref{sec:ablation}), demonstrating that our design effectively corrects over-confident semantic predictions.

\begin{figure*}[t]
    % \vspace{-0.3cm}
    \begin{center}
    \includegraphics[width=1\linewidth]{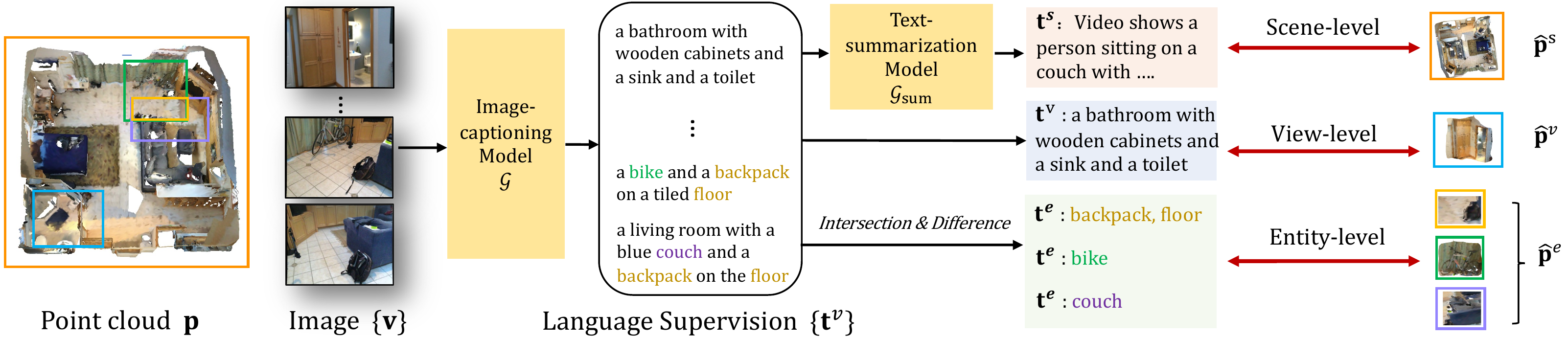}
    \end{center}
    \vspace{-0.5cm}
    % captionsetup{font={small}}
    \caption{Image-bridged point-language association. We present hierarchical scene-level, view-level and entity-level point-language association manners to assign partial point set with caption supervision through multi-view RGB images and VL foundation models.}
    \vspace{-0.3cm}
    \label{fig:caption}
    \end{figure*}

\subsection{Image-Bridged Point-Language Association}\label{sec:image-language}
\vspace{-0.1cm}
With the text-embedded classifier and the binary semantic calibration module, we obtain a deep model with open-vocabulary capability. 
Nevertheless, its performance on novel categories is very close to random guesses as shown in Table~\ref{tab:component}.  
Recent success of open-vocabulary works~\cite{li2022languagedriven,Hanoona2022Bridging,gu2021open} in 2D vision community shows the effectiveness of introducing language supervision to guide vision backbones.
Language supervision can not only enable the vision backbone to access abundant semantic concepts with a large vocabulary size but also assist in mapping vision and language features into a common space to facilitate multi-modality downstream tasks. 
However, Internet-scale paired point-text data are not as readily available as image-text pairs on social media, which largely hinders the development of language-driven 3D understanding.

To address this challenge, we propose PLA, an image-bridged point-language association module to provide language supervision for 3D scene perception without human annotation (see Fig.~\ref{fig:framework} \& Fig.~\ref{fig:caption}). 
Our core idea is to use multi-view images of a 3D scene as a bridge to access knowledge encoded in VL foundation models. 
As shown in Fig.~\ref{fig:caption}, a text description is first generated by a powerful image-captioning model taking images of 3D scenes as input, and then associated with a set of points in the 3D scene using the projection matrix between images and 3D scenes. We elaborate on our captioning procedure as well as the designed hierarchical point-caption association as follows.

\vspace{-0.3cm}
\subsubsection{Caption Multi-View Images}\label{sec:image-language}
\vspace{-0.1cm}

As image captioning is a fundamental task in VL research area~\cite{hossain2019comprehensive}, various foundation models~\cite{wang2022ofa,vit-gpt2,mokady2021clipcap} trained with massive samples are readily available for solving this task.
Specifically, taking the $j^\text{th}$ image of the $i^\text{th}$ scene $\textbf{v}_{ij}$ as input, the pre-trained image-captioning model $\mathcal{G}$ can generate its corresponding language description $\textbf{t}^v_{ij}$ as follows,
\vspace{-0.2cm}
\begin{equation}
\vspace{-0.2cm}
    \textbf{t}^v_{ij} = \mathcal{G}(\textbf{v}_{ij}). ~~
\end{equation}
Surprisingly, though $\mathcal{G}$ has not been specifically trained on the 3D scene understanding dataset, the entities in generated captions already cover the whole semantic label space of the popular 3D scene understanding dataset ScanNet~\cite{dai2017scannet}. In addition, the caption $\textbf{t}$
provides fairly accurate and comprehensive descriptions for room types, semantic categories with color and texture attributes, and even spatial relations (see language supervision $\{\mathbf{t}^v\}$ examples in Fig.~\ref{fig:caption} and more examples in Suppl.).

\vspace{-0.3cm}
\subsubsection{Associate Point Cloud with Language}\label{sec:3D-language}
\vspace{-0.1cm}
Given the image-caption pairs, the next step is to connect a point set $\mathbf{\hat{p}}$ to language $\mathbf{t}$ with images $\textbf{v}$ as bridge as follows:
\vspace{-0.25cm}
\begin{equation}
\vspace{-0.2cm}
    \text{Explore}~\langle\mathbf{\hat{p}},\mathbf{t}\rangle ~\text{with}~\langle\mathbf{\hat{p}, \mathbf{v}}\rangle ~\text{and}~\langle\mathbf{v},\mathbf{t}\rangle.
\end{equation}
Here, we propose three association fashions on point sets with different spatial scales.

\vspace{0.05in}\noindent\textbf{Scene-Level Point-Caption Association.} 
The simplest and coarsest association manner is to link language supervision to all points in a given 3D point cloud scene $\mathbf{\hat{p}}^s = \mathbf{p}$.  
As illustrated in Fig.~\ref{fig:caption}, we take all 2D image captions ${\mathbf{t}_{ij}^v}$ of a given scene $\mathbf{p}_j$ to obtain a scene-level caption $\mathbf{t}^s_j$ via a text summarizer~\cite{lewis2019bart} $\mathcal{G}_\text{sum}$ as follows:
\vspace{-0.2cm}
\begin{equation}
\vspace{-0.2cm}
    \mathbf{t}^s_j=\mathcal{G}_\text{sum}(\{\mathbf{t}_{1j}^v,\mathbf{t}_{2j}^v,\cdots \mathbf{t}_{n_jj}^v\}),
\end{equation}
where $n_j$ is the number of images of scene $\mathbf{p}_j$.
% where $\mathbf{t}^s_x$ is the scene-level caption for scene $\mathbf{p}_x$.
By forcing each scene $\mathbf{p}$ to learn from the corresponding scene descriptions $\mathbf{t}^s$, abundant vocabulary and visual-semantic relationships are introduced to improve the language understanding capability of a 3D network. Despite the simplicity of scene-level caption, we empirically find that it can lift the model's open-vocabulary capability by a large margin (see Sec.~\ref{sec:ablation}). 

\vspace{0.05in}
\noindent\textbf{View-Level Point-Caption Association.}
Albeit effective, scene-level caption only provides a single caption for all points in a scene, which overlooks the relation of language to local 3D point clouds, rendering it sub-optimal for scene understanding tasks. 
In this regard, we further propose a view-level point-caption association that leverages the geometrical relationship between image and points to assign each image caption $\mathbf{t}^v$ with a point set inside the 3D view frustum  ${\mathbf{\hat{p}}}^v$ of the given image $\mathbf{v}$ (see \textcolor{myblue}{blue} box in Fig.~\ref{fig:caption}). 
Specifically, to obtain the view-level point set $\mathbf{\hat{p}^v}$, we first back-project the RGB image $\mathbf{v}$ to 3D space using the depth information $\mathbf{d}$ to get its corresponding point set $\mathbf{\ddot{p}}$: 
\vspace{-0.2cm}
\begin{equation}\label{eq:projection}
\vspace{-0.2cm}
    \left[\begin{array}{c|c}
        \mathbf{\ddot{p}} & \mathbf{1}
    \end{array}\right]
     = \mathbf{T}^{-1}
      \left[\begin{array}{c|c}
        \mathbf{v} & \mathbf{d}
    \end{array}\right],
\end{equation}
where $[\cdot|\cdot]$ denotes block matrix,  
$\mathbf{T} \in \mathbbm{R}^{3\times4}$ is the projection matrix comprising of camera intrinsic matrix and rigid transformations obtained by sensor configurations or mature SLAM approaches~\cite{dai2017bundlefusion}.
As back-projected points $\mathbf{\ddot{p}}$ and points in 3D scene $\mathbf{p}$ may be only partially overlapped, we then compute their overlapped regions to get the view-level point set $\mathbf{\hat{p}}^v$ as follows, 
\vspace{-0.2cm}
\begin{equation}\label{eq:overlap}
\vspace{-0.2cm}
    \mathbf{\hat{p}}^v = V^{-1}(R(V(\mathbf{\ddot{p}}), V(\mathbf{p}))),
\end{equation}
where $V$ and $V^{-1}$ are the voxelization and reverse-voxelization processes, and $R$ denotes the radius-based nearest-neighbor search~\cite{Zhou2018}.
Such a view-based association enables the model to learn with region-level language description, which largely strengthens the model's recognition and localization ability on unseen categories. 

\vspace{0.05in}\noindent\textbf{Entity-Level Point-Caption Association.} 
Although view-level caption can already associate each image-caption $\mathbf{t}^v$ with a concrete partial point set in a 3D scene, such an association still constructs on a large 3D area (\ie around 25K points) with multiple semantic objects/categories as shown in Fig.~\ref{fig:caption}. 
This is not friendly for the 3D network to learn fine-grained point-wise semantic attributes and instance-wise position information from caption supervision.
In this regard, we further propose a fine-grained point-language association that owns the potential to build entity-level point-caption pairs, {\ie} object instances with a caption.

Specifically, as illustrated in Fig.~\ref{fig:caption}, we leverage the differences and intersections of adjacent view-level point sets $\mathbf{\hat{p}}^v$ and their corresponding view-caption $\mathbf{t}^v$ to obtain the entity-level associated points $\mathbf{\hat{p}}^e$ and caption $\mathbf{t}^e$.
First, we calculate entity-level caption $\mathbf{t}^e$ as below:
\vspace{-0.15cm}
\begin{small}
\begin{gather}\label{eq:diff_and_intersect}
\vspace{-0.15cm}
    w_i=E(\mathbf{t}_{i}^v), \\
    w^e_{i\setminus j}=w_i\setminus w_j,~~
    w^e_{j\setminus i}=w_j\setminus w_i,~~
    w^e_{i\cap j}=w_i\cap w_j,\\
    \mathbf{t}^e=\text{Concate}(w^e),
\end{gather}
\end{small}where $E$ denotes extracting a set of entity words $w$ from caption $\mathbf{t}^v$, 
% $i$ and $j$ index adjacent image views $\mathbf{v}$, 
$\setminus$ and $\cap$ represent the set difference and intersection separately, and Concate denotes the concatenation of all words with spaces to form an entity-level caption $\mathbf{t}^e$.
Similarly, we can easily calculate entity-level point sets and associate them to previously obtained entity-level captions to form point-caption pairs as below:
{\small{\setlength\abovedisplayskip{4.0pt}
\setlength\belowdisplayskip{4.1pt}
\begin{gather}\label{eq:diff_and_intersect}
    \mathbf{\hat{p}}^e_{i\setminus j} = (\mathbf{\hat{p}}^v_i\setminus \mathbf{\hat{p}}^v_j), ~~ \mathbf{\hat{p}}^e_{j\setminus i}=(\mathbf{\hat{p}}^v_j \setminus \mathbf{\hat{p}}^v_i),  ~~
    \mathbf{\hat{p}}^e_{i\cap j} = \mathbf{\hat{p}}^v_i \cap \mathbf{\hat{p}}^v_j,\\
    <\mathbf{\hat{p}}^e_{i\setminus j}, \mathbf{t}^e_{i\setminus j}>,<\mathbf{\hat{p}}^e_{j\setminus i}, \mathbf{t}^e_{j\setminus i}>, <\mathbf{\hat{p}}^e_{i\cap j}, \mathbf{t}^e_{i\cap j}>.
\end{gather}}}
With entity-level $\langle\mathbf{\hat{p}}^e$, $\mathbf{t}^e\rangle$ pairs, we further filter them to ensure  each entity-level points set $\mathbf{\hat{p}}^e$ relates to at least one entity and focuses on a small enough 3D space as follows,
\vspace{-0.15cm}
\begin{equation}\label{eq:filter}
\vspace{-0.15cm}
    \gamma < |\mathbf{\hat{p}}^e| < \delta\cdot\min(|\mathbf{\hat{p}}^v_{i}|, |\mathbf{\hat{p}}^v_{j}|) ~~\text{and}~~ |\mathbf{t}^e| > 0,
\end{equation}where $\gamma$ is a scalar to define minimal number of points, $\delta$ is a ratio to control the maximum size of $\mathbf{\hat{p}}^e$, and caption $\mathbf{t}^e$ is not empty. 
Such a constraint helps focus on a fine-grained 3D space with fewer entities in each caption supervision.

\begin{table}[htbp]
% \vspace{-0.2cm}
    \centering
    \begin{small}
    \setlength\tabcolsep{3pt}
    \scalebox{1.0}{
        \begin{tabular}{l|ccc}
            \bottomrule[1pt]
             & scene-level & view-level & entity-level\\
            \hline
            complexity & simplest & middle & hardest \\
            \hline
            \# captions & 1,201 & 24,902 & 6,163 \\
            \hline
            \# points for each caption & 145,171 & 24,294 & 3,933\\
            \toprule[0.8pt]
        \end{tabular}
    }
    \end{small}
    \vspace{-0.4cm}
    \caption{Comparison among point-caption association manners.}
    \vspace{-0.4cm}
    \label{tab:caption_statistics}
\end{table}

\vspace{0.05in}\noindent\textbf{Comparison Among Different Point-Caption Association Manners.}
The above-proposed three coarse-to-fine point-caption association manners actually hold different merits and drawbacks. As shown in Table~\ref{tab:caption_statistics}, the scene-level association has the simplest implementation but obtains the coarsest correspondence between captions and points (\ie each caption corresponds to over 140K points); the view-level association provides point-language mapping relation at a finer level, enjoying a larger semantic label space (\ie over 20$\times$ more captions) and a more localized point set (\ie around 6$\times$ fewer corresponding points per caption) than scene caption; 
the entity-level association owns the most fine-grained correspondence relation, matching each caption to only 4K points on average, and thus can further benefit dense prediction and instance localization in downstream tasks. 
We empirically show that the fine-grained association and the semantic-rich label space are two important factors for open-vocabulary perception tasks (see Sec.~\ref{sec:ablation}).

\subsection{Contrastive Point-Language Training}
~\label{sec:caption_formulation}
With obtained point-caption pairs $\langle\mathbf{\hat{p}}, \mathbf{t}\rangle$, we are ready to guide the 3D network F$_\text{3D}$ to learn from vocabulary-rich language supervisions. Here, we introduce a general point-language feature contrastive learning that can be applied to all kinds of coarse-to-fine point-caption pairs.

Specifically, we first obtain caption embeddings $\mathbf{f}^t$ with a pre-trained text encoder $\text{F}_{\text{text}}$. 
As for the associated partial point set $\mathbf{\hat{p}}$, we 
select its corresponding point-wise features from adapted features $\mathbf{f}^v$ and leverage global average pooling to obtain its feature vector $\mathbf{f}^{\hat{p}}$ as follows, 
\vspace{-0.15cm}
\begin{equation}\label{eq:caption_features}
\vspace{-0.1cm}
    \mathbf{f}^t = \text{F}_{\text{text}}(\mathbf{t}), ~~ \mathbf{f}^{\hat{p}} = \text{Pool}(\mathbf{\hat{p}}, \mathbf{f}^v).
\end{equation}
We then adopt contrastive loss as~\cite{zareian2021open} to pull corresponding point-caption feature embeddings closer and push away unrelated point-caption features as follows,
\vspace{-0.15cm}
\begin{equation}\label{eq:contrastive}
\vspace{-0.1cm}
    \mathcal{L}_{\text{cap}}=-\frac{1}{n_t}\sum\limits_{i=1}^{n_t} \log \frac{\exp(\mathbf{f}^{\hat{p}}_i\cdot\mathbf{f}^t_i/\tau)}{\sum_{j=1}^{n_t} \exp(\mathbf{f}^{\hat{p}}_i\cdot\mathbf{f}^t_j/\tau)},
\end{equation}
where $n_t$ is the number of point-caption pairs in any given association fashion and $\tau$ is a learnable temperature to modulate the logits as CLIP~\cite{radford2021learning}. 
It is also worth noting that we remove duplicate captions in a batch to avoid noisy optimization during contrastive learning. With Eq.~\eqref{eq:caption_features} and Eq.~\eqref{eq:contrastive}, we can easily compute caption losses on scene-level $\mathcal{L}_\text{cap}^s$, view-level $\mathcal{L}_\text{cap}^v$ and entity-level $\mathcal{L}_\text{cap}^e$.
Our final caption loss is a weighted combination as follows,
\vspace{-0.15cm}
\begin{equation}\label{eq:cap}
\vspace{-0.1cm}
\mathcal{L}^{\text{all}}_\text{cap}=\alpha_1*\mathcal{L}_\text{cap}^s+\alpha_2*\mathcal{L}_\text{cap}^v+\alpha_3*\mathcal{L}_\text{cap}^e,
\end{equation}
where $\alpha_1$, $\alpha_2$ and $\alpha_3$ are trade-off factors. As shown in Fig.~\ref{fig:framework}, the overall training objective can be written as
\vspace{-0.15cm}
\begin{equation}
\vspace{-0.15cm}
    \mathcal{L}=\mathcal{L}_{\text{sem}}+\mathcal{L}_{\text{loc}}+ \mathcal{L}_{\text{cap}}^{\text{all}}+\mathcal{L}_{\text{bi}}.
\end{equation}

%-------------------------------------------------------------------------

% sem seg
\begin{table*}[htbp]
    \vspace{-0.1cm}
    \centering
    \begin{small}
    \setlength\tabcolsep{2pt}
    \scalebox{0.92}{
        \begin{tabular}{c|c|ccc|ccc|ccc|ccc|ccc}
            \bottomrule[1pt]
            \multirow{3}{*}{Method} & \multirow{3}{*}{$\mathcal{C}^N$ prior}  & \multicolumn{9}{c|}{ScanNet} & \multicolumn{6}{c}{S3DIS} \\
            \cline{3-17}
            & & \multicolumn{3}{c|}{B15/N4} & \multicolumn{3}{c|}{B12/N7} & \multicolumn{3}{c|}{B10/N9} & \multicolumn{3}{c|}{B8/N4} & \multicolumn{3}{c}{B6/N6} \\
            \cline{3-17}
            & & hIoU & mIoU$^\mathcal{B}$ & mIoU$^\mathcal{N}$ & hIoU & mIoU$^\mathcal{B}$ & mIoU$^\mathcal{N}$ & hIoU & mIoU$^\mathcal{B}$ & mIoU$^\mathcal{N}$ & hIoU & mIoU$^\mathcal{B}$ & mIoU$^\mathcal{N}$ & hIoU & mIoU$^\mathcal{B}$ & mIoU$^\mathcal{N}$ \\
            \hline
            LSeg-3D~\cite{li2022languagedriven} & $\times$ & 00.0 & 64.4 & 00.0 & 00.9 & 55.7 & 00.1 & 01.8 & 68.4 & 00.9 & 00.1 & 49.0 & 00.1 & 00.0 & 30.1 & 00.0 \\ 
            3DGenZ~\cite{michele2021generative} & \checkmark & 20.6 & 56.0 & 12.6 & 19.8 & 35.5 & 13.3 & 12.0 & 63.6 & 06.6 & 08.8 & 50.3 & 04.8 & 09.4 & 20.3 & 06.1 \\
            3DTZSL~\cite{cheraghian2020transductive} & \checkmark & 10.5 & 36.7 & 06.1 & 03.8 & 36.6 & 02.0 & 07.8 & 55.5 & 04.2 & 08.4 & 43.1 & 04.7 & 03.5 & 28.2 & 01.9 \\
            \hline
            PLA (w/o Cap.) & $\times$  & 39.7 & \textbf{68.3} & 28.0 & 24.5 & \textbf{70.0} & 14.8 & 25.7 & 75.6 & 15.5 & 13.0 & 58.0 & 07.4 & 12.2 & 54.5 & 06.8\\
            PLA & $\times$ & \textbf{65.3} & \textbf{68.3} & \textbf{62.4} & \textbf{55.3} & 69.5 & \textbf{45.9} & \textbf{53.1} & \textbf{76.2} & \textbf{40.8} & \textbf{34.6} & \textbf{59.0} & \textbf{24.5} & \textbf{38.5} & \textbf{55.5} & \textbf{29.4} \\
            PLA (w/ self-train) &\checkmark & 70.3 & 68.9 & 71.7 & 61.1 & 70.4 & 54.0 & 59.2 & 76.9 & 48.2 & 36.1 & 59.7 & 26.0 & 46.7 & 58.9 & 38.7 \\
            \hline
            Fully-Sup. & \checkmark & 73.3 & 68.4 & 79.1 & 70.6 & 70.0 & 71.8 & 69.9 & 75.8 & 64.9 & 67.5 & 61.4 & 75.0 & 65.4 & 59.9 & 72.0 \\
            \toprule[0.8pt]
        \end{tabular}
     }
    \end{small}
    \vspace{-0.4cm}
    \caption{Results for open-vocabulary 3D semantic segmentation on ScanNet and S3DIS in terms of hIoU, mIoU$^\mathcal{B}$ and mIoU$^\mathcal{N}$. $\mathcal{C}^N$ prior denotes whether novel category names $\mathcal{C}^N$ need to be known during training. PLA (w/o Cap.) denotes training without point-caption pairs as supervision. Best open-vocabulary results are highlighted in \textbf{bold}.}
    \label{tab:sem_seg}
    \vspace{-0.2cm}
\end{table*}

% inst seg
\begin{table*}[htbp]
    \centering
    \begin{small}
    \setlength\tabcolsep{2pt}
    \scalebox{0.9}{
        \begin{tabular}{c|c|ccc|ccc|ccc|ccc|ccc}
            \bottomrule[1pt]
            \multirow{3}{*}{Method} & \multirow{3}{*}{$\mathcal{C}^N$ prior} & \multicolumn{9}{c|}{ScanNet} & \multicolumn{6}{c}{S3DIS} \\
            \cline{3-17}
            & & \multicolumn{3}{c|}{B13/N4} & \multicolumn{3}{c|}{B10/N7} & \multicolumn{3}{c|}{B8/N9} & \multicolumn{3}{c|}{B8/N4} & \multicolumn{3}{c}{B6/N6} \\
            \cline{3-17}
            & & hAP$_{50}$ & mAP$_{50}^\mathcal{B}$ & mAP$_{50}^\mathcal{N}$ & hAP$_{50}$ & mAP$_{50}^\mathcal{B}$ & mAP$_{50}^\mathcal{N}$ & hAP$_{50}$ & mAP$_{50}^\mathcal{B}$ & mAP$_{50}^\mathcal{N}$ & hAP$_{50}$ & mAP$_{50}^\mathcal{B}$ & mAP$_{50}^\mathcal{N}$ & hAP$_{50}$ & mAP$_{50}^\mathcal{B}$ & mAP$_{50}^\mathcal{N}$ \\
            \hline
            LSeg-3D~\cite{li2022languagedriven} & $\times$ & 05.1 & 57.9 & 02.6 & 02.0 & 50.7 & 01.0 & 02.4 & 59.4 & 01.2 & 00.5 & 58.3 & 00.3 & 01.1 & 41.4 & 00.5 \\ 
            \hline
            PLA (w/o Cap.) & $\times$ & 21.0 & \textbf{59.6} & 12.6 & 11.1 & \textbf{56.2} & 06.2 & 15.9 & \textbf{63.2} & 09.1 & 01.8 & \textbf{59.3} & 00.9 & 01.3 & \textbf{49.2} & 01.2 \\
            PLA & $\times$ & \textbf{55.5} & 58.5 & \textbf{52.9} & \textbf{31.2} & 54.6 & \textbf{21.9} & \textbf{35.9} & 63.1 & \textbf{25.1} & \textbf{15.0} & 59.0 & \textbf{08.6} & \textbf{16.0} & 46.9 & \textbf{09.8} \\
            PLA (w/ self-train) & \checkmark & 58.6 & 58.0 & 59.2 & 41.4 & 56.9 & 32.6 & 42.1 & 61.1 & 32.1 & 26.7 & 60.3 & 17.2 & 23.4 & 45.6 & 15.8 \\
            \hline
            Fully-Sup. & \checkmark & 64.5 & 59.4 & 70.5 & 62.5 & 57.6 & 62.0 & 62.0 & 65.1 & 62.0 & 57.6 & 60.8 & 54.6 & 57.4 & 50.0 & 67.5 \\
            \toprule[0.8pt]
        \end{tabular}
     }
    \end{small}
    \vspace{-0.4cm}
    \caption{Results for open-vocabulary 3D instance segmentation on ScanNet and S3DIS in terms of hAP$_{50}$, mAP$_{50}^\mathcal{B}$ and mAP$_{50}^\mathcal{N}$.}
    \vspace{-0.45cm}
    \label{tab:inst_seg}
\end{table*}

\section{Experiments}
\vspace{-0.1cm}
\subsection{Basic Setups}
\vspace{-0.2cm}
\noindent\textbf{Datasets and Perception Tasks.}
To validate the effectiveness of our point-language association paradigm, we conduct experiments on two datasets: ScanNet~\cite{dai2017scannet} densely annotated in 20 classes and S3DIS~\cite{armeni20163d} with 13 classes on both semantic and instance segmentation tasks.

\vspace{0.05in}
\noindent\textbf{Category Partitions.}
Without standard open-vocabulary partitions on these two datasets, we build an open-vocabulary benchmark with multiple base/novel partitions. 
To circumvent model confusion, we disregard the ``otherfurniture'' class in ScanNet and the ``clutter'' class in S3DIS as they lack exact semantic meanings and can include any semantic categories.
As for ScanNet, we randomly partition the rest 19 classes into 3 base/novel partitions for semantic segmentation, \ie B15/N4, B12/N7 and B10/N9, where B15/N4 indicates 15 base and 4 novel categories. We also follow SoftGroup~\cite{vu2022softgroup} to exclude two background classes and thus obtain B13/N4, B10/N7, and B8/N9 partitions for instance segmentation on ScanNet.
As for S3DIS, we randomly shuffle the rest 12 classes into 2 base/novel splits, \ie B8/N4, B6/N6 for both semantic and instance segmentation. Specific category splits are presented in the Suppl..

\vspace{0.05in}
\noindent\textbf{Metrics.} 
We employ widely adopted mean intersection over union (mIoU) and mean average precision under 50\% IoU threshold (mAP$_{50}$) as evaluation metrics for semantic and instance segmentation, respectively. 
These metrics are calculated on base and novel classes separately with superscripts of $\mathcal{B}$ and $\mathcal{N}$ (\eg mIoU$^\mathcal{B}$). 
Further, we use harmonic mean IoU (hIoU) and AP$_{50}$ (hAP$_{50}$) as major indicators following popular zero-shot learning works~\cite{xian2019semantic,xu2021simple} to consider category partition between base and novel.

\vspace{0.05in}
\noindent\textbf{Architectures and Baseline Methods.} We adopt the popular and high-performance sparse convolutional UNet~\cite{graham20183d,choy20194d} as 3D encoder $\text{F}_{\text{3D}}$, the text encoder of CLIP as $\text{F}_{\text{text}}$, two fully-connected layers with batch normalization~\cite{ioffe2015batch} and ReLU~\cite{nair2010rectified} as VL adapter $\text{F}_{\theta}$, an UNet decoder as binary head $\text{F}_{\text{b}}$. Also, we utilize the state-of-the-art instance segmentation network SoftGroup~\cite{vu2022softgroup} for instance head F$_\text{ins}$.

As for baseline methods, other than the above-mentioned \textbf{LSeg-3D} in Sec.\ref{sec:text_embed_cls}, we also re-produce two 3D zero-shot learning methods \textbf{3DGenZ}~\cite{michele2021generative} and \textbf{3DTZSL}~\cite{cheraghian2020transductive} with task-tailored modifications. The implementation details are provided in the Suppl..

\vspace{-0.1cm}
\subsection{Main Results}\label{sec:sem_seg}
\vspace{-0.2cm}
\noindent\textbf{3D Semantic Segmentation.}
As shown in Table~\ref{tab:sem_seg}, compared to LSeg-3D~\cite{li2022languagedriven} baseline, our method  obtains around 51.3\% $\sim$ 65.3\% and 34.5\% $\sim$ 38.5\% hIoU improvements among different partitions on ScanNet and S3DIS respectively, demonstrating its superior open-vocabulary capability.
Even compared to previous zero-shot methods 3DGenZ~\cite{michele2021generative} and 3DTZSL~\cite{cheraghian2020transductive} that know novel category names during training, our method still obtains 35.5\% $\sim$ 54.8\% improvements in terms of hIoU among various partitions on ScanNet. 
Especially, our PLA trained model largely surpasses its no caption supervision counterparts (\ie PLA (w/o Cap.)) by 25.6\% $\sim$ 30.8\% hIoU and 21.6\% $\sim$ 26.3\% hIoU on ScanNet and S3DIS, respectively.
It is noteworthy that the improvement from our method is consistent on different base/novel partitions and datasets, further illustrating its robustness and effectiveness.

\vspace{0.05in}
\noindent\textbf{3D Instance Segmentation.}
As demonstrated in Table~\ref{tab:inst_seg}, our method remarkably surpasses baseline methods by 29.2\% $\sim$ 50.4\% hAP$_{50}$ and 14.5\% $\sim$ 14.9\% hAP$_{50}$ among different base/novel partitions on ScanNet and S3DIS, respectively. 
Such outstanding performance indicates our contrastive point-language training helps the 3D backbone learn not only semantic attributes but also 
instance localization information from captions.
Notice that the improvement for S3DIS is slighter than ScanNet on both semantic segmentation and instance segmentation. This is actually caused by S3DIS's small number of training samples (only 271 scenes) and much fewer point-caption pairs owing to fewer overlapped regions between images and 3D scenes.

\vspace{0.05in}
\noindent\textbf{Self-Bootstrap with Novel Category Prior.} As some existing zero-shot methods (\ie 3DGenZ~\cite{michele2021generative} and 3DTZSL~\cite{cheraghian2020transductive}) can access novel category names but no human-annotation during training, here we also provide a 
simple variant to leverage such novel category prior in self-training fashion~\cite{xie2020self}. As shown in Table~\ref{tab:sem_seg} and~\ref{tab:inst_seg}, PLA (w/ self-train) obtains around 2\% $\sim$ 12\% gains among semantic and instance segmentation on two datasets. This demonstrates that our model can further self-bootstrap its zero-shot capability and extend its vocabulary size without any human annotation. 

\vspace{-0.1cm}
\subsection{Zero-shot Domain Transfer}\label{sec:transfer}
\vspace{-0.2cm}
Our method already shows excellent potential in solving in-domain open-vocabulary scene understanding tasks with category shifts.
However, transferable open-vocabulary learners across different domains/datasets also merit exploration, as they face both category and data distribution shifts.
In this regard, we conduct zero-shot domain transfer experiments that train the model on ScanNet's base classes and test it on all S3DIS classes without fine-tuning. 
Notably, S3DIS has 4 categories not present in ScanNet.
As shown in Table~\ref{tab:transfer}, our PLA consistently outperforms LSeg-3D~\cite{li2022languagedriven} by $7.7\%\sim 18.3\%$ mIoU for semantic segmentation and $5.0\% \sim 9.5\%$ mAP$_{50}$ for instance segmentation.
Such outstanding improvements substantiate our model's generality for both category shift and data distribution shift.
Note that we do not use the binary head for domain transfer here, as the base/novel partition is dataset-specific.  
We leave calibrating base and novel semantic predictions in out-of-domain open-vocabulary scenarios to future work. 

\vspace{-0.2cm}
\section{Ablation Studies}\label{sec:ablation}
\vspace{-0.1cm}
In this section, we examine key components of our framework through in-depth ablation studies. Experiments are conducted on ScanNet B15/N4 partition by default. The default setting is marked in {\colorbox{mygray}{gray}}.

\begin{table}[htbp]
    \vspace{-0.3cm}
    \centering
    \begin{small}
    \setlength\tabcolsep{6pt}
    \scalebox{0.95}{
        \begin{tabular}{c|c|c|c|c}
            \bottomrule[1pt]
            \multirow{2}{*}{\makecell[c]{ScanNet \\ partition}} & \multicolumn{2}{c|}{S3DIS Semantic (mIoU)} & \multicolumn{2}{c}{{S3DIS Instance (mAP$_{50}$)}} \\
            \cline{2-5}
            & LSeg-3D & PLA & LSeg-3D & PLA \\
            \hline
            B19/N0 & 42.5 & 50.2~{\color{blue}{(+7.7)}} & 37.5 & 43.6~{\color{blue}{(+6.1)}} \\
            \hline
            B15/N4 & 30.2 & 48.5~{\color{blue}{(+18.3)}} & 31.2 & 40.7~{\color{blue}{(+9.5)}} \\
            \hline
            B12/N7 & 26.1 & 38.3~{\color{blue}{(+12.2)}} & 28.2 & 35.1~{\color{blue}{(+6.9)}} \\
            \hline
            B10/N9 & 34.5 & 48.1~{\color{blue}{(+13.6)}} & 33.8 & 38.8~{\color{blue}{(+5.0)}} \\
            % \hline
            % / & Fully-Sup. & 65.9 & 58.8 \\
            \toprule[0.8pt]
        \end{tabular}
    }
    \end{small}
    \vspace{-0.3cm}
    \caption{Zero-shot domain transfer results for semantic segmentation and instance segmentation on ScanNet $\rightarrow$ S3DIS.}
    \vspace{-0.5cm}
    \label{tab:transfer}
\end{table}

\vspace{0.05in}
\noindent\textbf{Component Analysis.}
We investigate the effectiveness of our proposed binary calibration module and three coarse-to-fine point-caption supervision here.
As shown in Table~\ref{tab:component}, adopting binary head for semantic calibration greatly surpasses baseline LSeg-3D by $39.8\%$ hIoU on semantic segmentation and $15.9\%$ hAP$_{50}$ on instance segmentation. 
Such performance lifts on both base and novel classes verify that it correctly rectifies semantic scores.

As for point-caption association manners, they all substantially improve results by a large margin of 14.8\% $\sim$ 23.8\% hIoU and 31.8\% $\sim$ 35.6\% hAP$_{50}$ on semantic and instance segmentation, respectively.
Among three association fashions, entity-level caption supervision performs the best, demonstrating that fine-grained language-point correspondence is one of the most vital considerations for constructing point-caption pairs.
Notice that when we combine different types of captions, the model will not always obtain improvements in all scenarios, potentially caused by the difficulty of simultaneously optimizing multiple caption losses with various granularities on some tasks.

\begin{table}[htbp]
    \vspace{-0.3cm}
    \centering
    \begin{small}
    \setlength\tabcolsep{1pt}
    \scalebox{0.85}{
        \begin{tabular}{c|c|c|c|c|c}
            \bottomrule[1pt]
            \multicolumn{4}{c|}{Components} & \multirow{2}{*}{hIoU / mIoU$^\mathcal{B}$ /mIoU$^\mathcal{N}$} & \multirow{2}{*}{hAP$_{50}$ / mAP$_{50}^\mathcal{B}$ / mAP$_{50}^\mathcal{N}$} \\
            \cline{1-4}
            Binary & Cap$^s$ & Cap$^v$ & Cap$^e$ & & \\
            \hline
            & & & & 00.0 / 64.4 / 00.0 & 05.1 / 57.9 / 02.6\\
            \hline
            \checkmark & & & & 39.8 / \textbf{68.5} / 28.1 & 21.0 / \textbf{59.6} / 12.8 \\
            \hline
            \checkmark & \checkmark & & & 54.6 / 67.9 / 45.7 & 52.8 / 57.8 / 36.6 \\
            \checkmark & & \checkmark & & 61.3 / \textbf{68.5} / 55.5 & 55.9 / 58.9 / 53.3 \\
            \checkmark & & & \checkmark & 63.6 / 67.8 / 60.0 &  \textbf{56.6} / 59.0 / \textbf{54.4} \\
            \hline
            \checkmark & \checkmark & \checkmark & & 61.9 / 68.1 / 56.8 & 54.9 / 59.5 / 51.0 \\
            \checkmark &  & \checkmark & \checkmark & \cellcolor{mygray}\textbf{65.3} / 68.3 / \textbf{62.4} & \cellcolor{mygray}55.5 / 58.5 / 52.9 \\
            \checkmark & \checkmark & \checkmark & \checkmark & 64.6 / 69.0 / 60.8 & 54.5 / 58.2 / 51.4 \\
            \toprule[0.8pt]
        \end{tabular}
    }
    \end{small}
    \vspace{-0.3cm}
    \caption{Component analysis on ScanNet. Binary denotes binary head calibration. Cap$^s$, Cap$^v$ and Cap$^e$ denotes scene-level, view-level and entity-level caption supervision, respectively.}
    \vspace{-0.3cm}
    \label{tab:component}
\end{table}

% visualization results
\begin{figure*}[htbp]
    \vspace{-0.2cm}
    \begin{center}
    \scalebox{1.0}{
        \includegraphics[width=1\linewidth]{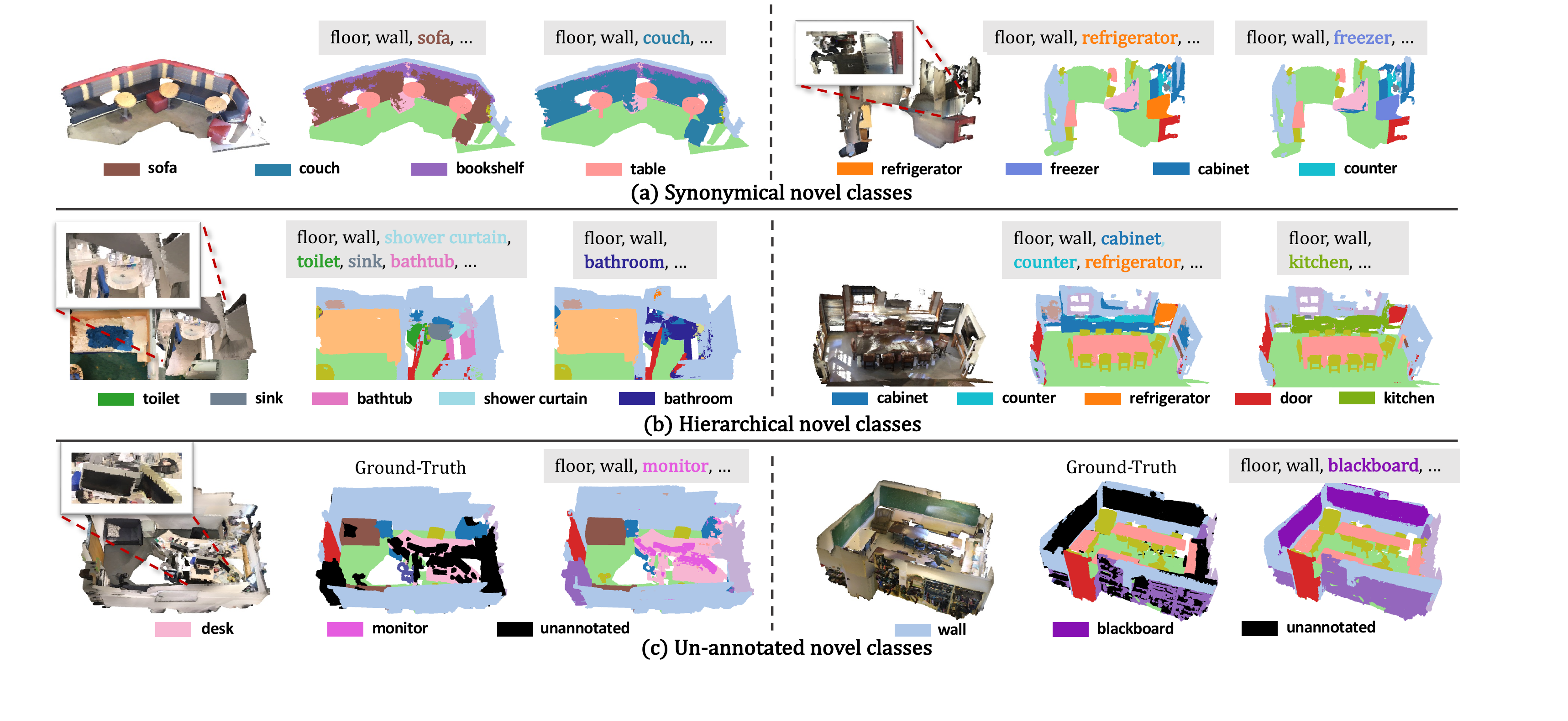}
    }
    \end{center}
    \vspace{-0.65cm}
    % captionsetup{font={small}}
    \caption{Qualitative results of recognizing out-of-vocabulary classes. (a) demonstrates the results of recognizing synonymical classes. (b) shows the segmentation results on abstract concepts. (c) presents the results of segmenting unannotated categories in the  dataset.}
    \vspace{-0.4cm}
    \label{fig:vis}
\end{figure*}

\noindent\textbf{Caption Composition Analysis.} As a caption can composite entities (\eg sofa), their relationships (\eg spatial relation) and attributes (\eg color and texture), we investigate which types of words mainly contribute to the open-vocabulary capability. As shown in Table~\ref{tab:caption_components}, when only keeping entity phrases in the caption, (a) variant even outperforms the full caption variant. In addition, if we only keep entities that exactly match category names in captions, obtained (b) variant suffers over $13\%$ mIoU degradation on novel categories, showing that diverse entity words to expand semantic space is a crucial factor for captions.
Furthermore, although the (c) variant introduces both correct base and novel label names in the caption, it still obtains slightly inferior performance to our foundation-model-generated caption, illustrating existing foundation models are powerful enough to provide promising supervision.

\begin{table}[htbp]
    \vspace{-0.3cm}
    \centering
    \begin{small}
    \setlength\tabcolsep{5pt}
    \scalebox{1.0}{
        \begin{tabular}{l|c}
            \bottomrule[1pt]
            Caption Composition & hIoU / mIoU$^\mathcal{B}$ / mIoU$^\mathcal{N}$ \\
            \hline
            (a) keep only entities & \textbf{65.7} / \textbf{69.0} / \textbf{62.7} \\
            (b) keep only label names & 57.6 / 68.5 / 49.6 \\
            (c) ground-truth label names & 64.8 / 68.1 / 61.9 \\
            (d) full caption & \cellcolor{mygray}65.3 / 68.3 / 62.4\\ 
            \toprule[0.8pt]
        \end{tabular}
    }
    \end{small}
    \vspace{-0.3cm}
    \caption{Ablation of caption composition.}
    \vspace{-0.5cm}
    \label{tab:caption_components}
\end{table}

\vspace{0.05in}
\noindent\textbf{Text Encoder Selection.}
Here, we compare different text encoders $\text{F}_\text{text}$ for extracting caption and category embeddings. As shown in Table~\ref{tab:text_encoder}, the vision-language pre-trained text encoder of CLIP~\cite{radford2021learning} shows over $7\%$ higher mIoU$^\mathcal{N}$ than BERT~\cite{devlin2018bert} and GPT2~\cite{radford2019language} that are only pre-trained on language modality.  
This demonstrates that the vision-aware text encoder can provide better language embedding for 3D-language tasks since 3D also leverages texture, shape and RGB information as images for recognition.

\begin{table}[htbp]
\vspace{-0.3cm}
    \centering
    \begin{small}
    \setlength\tabcolsep{2pt}
    \scalebox{0.82}{
        \begin{tabular}{c|c|c|c}
            \bottomrule[1pt]
            Text Encoder & BERT~\cite{devlin2018bert} & GPT2~\cite{radford2019language} & CLIP~\cite{radford2021learning} \\
            \hline
            hIoU / mIoU$^\mathcal{B}$ / mIoU$^\mathcal{N}$ & 61.2 / 68.7 / 55.2 & 61.0 / \textbf{69.1} / 54.6 & \cellcolor{mygray}\textbf{65.3} / 68.3 / \textbf{62.4}\\
            \toprule[0.8pt]
        \end{tabular}
    }
    \end{small}
    \vspace{-0.4cm}
    \caption{Ablation of text encoder.}
    \vspace{-0.4cm}
    \label{tab:text_encoder}
\end{table}

\noindent\textbf{Foundation Model for Image Captioning.}
By default, we employ one of the most popular open-source image captioning models, GPT-ViT2~\cite{vit-gpt2}, on the HuggingFace platform to generate captions in main experiments. 
However, as shown in Table~\ref{tab:foundation_model}, the recent state-of-the-art foundation model OFA~\cite{wang2022ofa} can consistently surpass GPT-ViT2 on three partitions, which reflects the potential of our method to be further boosted with stronger foundation models.

\begin{table}[htbp]
\vspace{-0.3cm}
    \centering
    \begin{small}
    \setlength\tabcolsep{3pt}
    \scalebox{0.9}{
        \begin{tabular}{l|c|c|c}
            \bottomrule[1pt]
            \multirow{2}{*}{model} & \multicolumn{3}{c}{hIoU / mIoU$^\mathcal{B}$ / mIoU$^\mathcal{N}$} \\
            \cline{2-4}
            & B15/N4 & B12/N7 & B10/N9 \\
            \hline
            ViT-GPT2~\cite{vit-gpt2} & \cellcolor{mygray}65.3 / \textbf{68.3} / 62.4 & \cellcolor{mygray}55.3 / 69.5 / 45.9 & \cellcolor{mygray}53.1 / \textbf{76.2} / 40.8 \\
            OFA~\cite{wang2022ofa} & \textbf{65.6} / \textbf{68.3} / \textbf{63.1} & \textbf{57.5} / \textbf{69.8} / \textbf{48.9} & \textbf{56.6} / 75.9 / \textbf{45.1} \\
            \toprule[0.8pt]
        \end{tabular}
    }
    \end{small}
    \vspace{-0.3cm}
    \caption{Ablation of VL foundation model for image captioning.}
    \vspace{-0.5cm}
    \label{tab:foundation_model}
\end{table}

\section{Qualitative Analysis}
\vspace{-0.1cm}

To more straightforwardly illustrate the open-vocabulary ability of our method, we present some interesting qualitative results in terms of recognizing synonymical classes, abstract classes and even unannotated classes.

\noindent\textbf{Synonymical Novel Classes.} 
Here, we substitute class names with related but new words for inference.
As illustrated in Fig.~\ref{fig:vis} (a), when we replace ``sofa'' with ``couch'' or ``refrigerator'' with ``freezer'', the model still attains a high-quality segmentation mask.
This demonstrates our model is robust to recognize synonymical concepts.

\noindent\textbf{Abstract Novel Classes.}
Apart from object entities, we find the model is able to understand more abstract concepts such as room types. 
As shown in Fig.~\ref{fig:vis} (b), by removing ``shower curtain'', ``toilet'', ``sink'' and ``bathtub'' in input categories and adding ``bathroom'', the predicted ``bathroom'' roughly covers the real bathroom region. The right example shows the model can also understand `kitchen' regions.
%despite ``refrigerator'' partly recognized as ``door''. 
It indicates our model is capable to recognize out-of-vocabulary and abstract concepts beyond concrete semantic objects.

\noindent\textbf{Unannotated Novel Classes.}
As current 3D datasets fail to annotate all classes due to insufferable annotation costs, our model owns the potential to recognize those unannotated classes with high-quality predictions, facilitating open-world applications.
As shown in Fig.~\ref{fig:vis} (c), the model successfully identifies ``monitor'' and ``blackboard'' that are not included in the dataset annotations with accurate masks. 

\vspace{-0.15cm}
\section{Conclusion}
\vspace{-0.2cm}
We propose PLA, a general and effective language-driven 3D scene understanding framework
that enables the 3D model to localize and recognize novel categories. By leveraging images as a bridge, we construct hierarchical point-language pairs harvesting powerful 2D VL foundation models and geometric constraints between 3D scenes and 2D images. We employ contrastive learning to pull features of such associated pairs closer, introducing rich semantic concepts into the 3D network. 
Extensive experimental results show the superiority of our method on not only in-domain open-vocabulary semantic and instance segmentation, but also challenging out-of-domain zero-shot transfer.

\vspace{0.1cm}
\noindent\textbf{Acknowledgement.} This work has been supported by Hong Kong Research Grant Council - Early Career Scheme (Grant No. 27209621), General Research Fund Scheme (Grant no. 17202422), and RGC matching fund scheme (RMGS). Part of the described research work is conducted in the JC STEM Lab of Robotics for Soft Materials funded by The Hong Kong Jockey Club Charities Trust.

%%%%%%%%% REFERENCES
{\small
\bibliographystyle{ieee_fullname}
\bibliography{egbib}
}

%%%%%%%%% SUPPL
\clearpage
% \newpage
\appendix

\crefname{section}{Sec.}{Secs.}
\Crefname{section}{Section}{Sections}
\Crefname{table}{Table}{Tables}
\crefname{table}{Tab.}{Tabs.}
\renewcommand{\thesection}{S\arabic{section}}
\renewcommand{\thetable}{S\arabic{table}}
\renewcommand{\thefigure}{S\arabic{figure}}

\centerline{\large{\textbf{Outline}}}
\vspace{0.3cm}
In this supplementary file, we provide more experimental results and details not elaborated on in our main paper due to page length limits:
\begin{itemize}
    \item Sec.~\ref{sec:implementation}: Details of our open-vocabulary scene understanding benchmark.
    \item Sec.~\ref{sec:analysis}: Limitation analysis of PointCLIP for scene understanding tasks.
    \item Sec.~\ref{sec:result}: Additional experimental results on re-partition results, per-class results, error bar results, fully-supervised results with caption supervision and combination of caption supervisions. 
    \item Sec.~\ref{sec:example}: Examples of image-caption pairs and hierarchical point-caption pairs.
    \item Sec.~\ref{sec:vis}: Qualitative results of open-vocabulary scene understanding.
    \item Sec.~\ref{sec:limitation}: Limitation and open problems.
\end{itemize}

\section{Implementation Details}\label{sec:implementation}
Here, we present the implementation details of dataset category partition, network modifications, baseline setups, hyper-parameter configurations and usage of images.

\subsection{Dataset Category Partition}

\begin{table*}[htbp]
    % \vspace{0.1cm}
    \centering
    \begin{small}
    \setlength\tabcolsep{5pt}
    \scalebox{1.0}{
        \begin{tabular}{l|l|l}
            \bottomrule[1pt]
            Partition & Base Categories & Novel Categories \\
            \hline
            B15/N4 & \makecell[l]{wall, floor, cabinet, bed, chair, table, door, window, picture,\\ counter, curtain, refrigerator, showercurtain, sink, bathtub} & sofa, bookshelf, desk, toilet\\
            \hline
            B12/N7 & \makecell[l]{wall, floor, cabinet, sofa, door, window, counter, desk,\\curtain, refrigerator, showercurtain, toilet} & bed, chair, table, bookshelf, picture, sink, bathtub \\
            \hline
            B10/N9 & \makecell[l]{wall, floor, cabinet, bed, chair, sofa, table, door, window,\\curtain} & \makecell[l]{bookshelf, picture, counter, desk, refrigerator, showercurtain,\\toilet, sink, bathtub}\\
            \toprule[0.8pt]
        \end{tabular}
    }
    \end{small}
    % \vspace{-0.3cm}
    \caption{Category partitions for open-vocabulary semantic segmentation on ScanNet.}
    % \vspace{-0.5cm}
    \label{tab:scannet_sem_partition}
\end{table*}

\begin{table*}[htbp]
    % \vspace{0.1cm}
    \centering
    \begin{small}
    \setlength\tabcolsep{5pt}
    \scalebox{1.0}{
        \begin{tabular}{l|l|l}
            \bottomrule[1pt]
            Partition & Base Categories & Novel Categories \\
            \hline
            B13/N4 & \makecell[l]{cabinet, bed, chair, table, door, window, picture, \\ counter, curtain, refrigerator, showercurtain, sink, bathtub} & sofa, bookshelf, desk, toilet\\
            \hline
            B10/N7 & \makecell[l]{cabinet, sofa, door, window, counter, desk, curtain, \\refrigerator, showercurtain, toilet} & bed, chair, table, bookshelf, picture, sink, bathtub \\
            \hline
            B8/N9 & \makecell[l]{cabinet, bed, chair, sofa, table, door, window, curtain} & \makecell[l]{bookshelf, picture, counter, desk, refrigerator, showercurtain, \\toilet, sink, bathtub}\\
            \toprule[0.8pt]
        \end{tabular}
    }
    \end{small}
    % \vspace{-0.3cm}
    \caption{Category partitions for open-vocabulary instance segmentation on ScanNet.}
    \label{tab:scannet_inst_partition}
\end{table*}

\begin{table*}[htbp]
    % \vspace{0.1cm}
    \centering
    \begin{small}
    \setlength\tabcolsep{8pt}
    \scalebox{1.0}{
        \begin{tabular}{l|l|l}
            \bottomrule[1pt]
            Partition & Base Categories & Novel Categories \\
            \hline
            B8/N4 & \makecell[l]{ceiling, floor, wall, beam, column, door, chair, board} & window, table, sofa, bookcase \\
            \hline
            B6/N6 & \makecell[l]{ceiling, wall, beam, column, chair, bookcase} &  \makecell[l]{floor, window, door, table, sofa, board} \\
            \toprule[0.8pt]
        \end{tabular}
    }
    \end{small}
    % \vspace{-0.3cm}
    \caption{Category partitions for open-vocabulary semantic and instance segmentation on S3DIS.}
    % \vspace{-0.5cm}
    \label{tab:s3dis_partition}
\end{table*}

As mentioned in Sec. {\color{red}4.1} of the main paper, we build a 3D open-vocabulary benchmark on ScanNet~\cite{dai2017scannet} and S3DIS~\cite{armeni20163d} with multiple base/novel partitions. ScanNet~\cite{dai2017scannet} consists of 1,613 scenes (1,201 scenes for training, 312 scenes for validation and 100 for testing) densely annotated in 20 classes. We discard the `otherfurniture' class and partition the rest 19 classes into three partitions for semantic segmentation as shown in Table~\ref{tab:scannet_sem_partition}. Note that the B15/N4 partition adheres to the 3DGenZ~\cite{michele2021generative} partitioning scheme. As for instance segmentation, we follow SoftGroup~\cite{vu2022softgroup} to ignore two background classes (\ie wall and floor) and obtain corresponding partitions (see Table~\ref{tab:scannet_inst_partition}).

S3DIS~\cite{armeni20163d} contains 271 scans across 6 building areas along with 13 categories. Following previous work~\cite{qi2017pointnet++}, we treat the 5th area as the validation split and other areas as the training split. We discard the `clutter' class and partition the rest 12 classes into two partitions for both semantic segmentation and instance segmentation as demonstrated in Table~\ref{tab:s3dis_partition}.

% network configuration
\subsection{Network Modifications}

\begin{figure*}[htbp]
    % \vspace{-0.3cm}
    \begin{center}
    \includegraphics[width=1\linewidth]{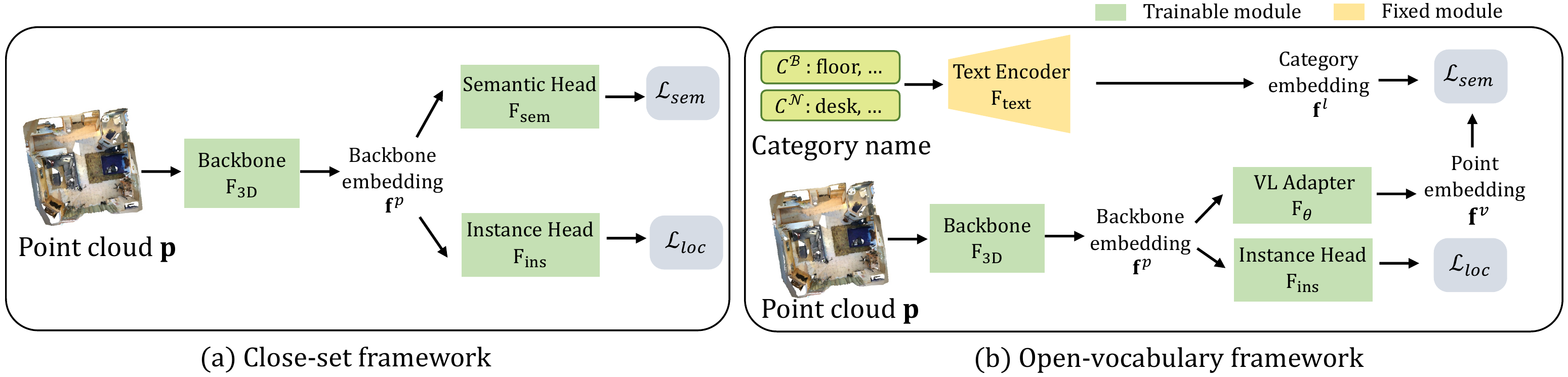}
    \end{center}
    % \vspace{-0.5cm}
    % captionsetup{font={small}}
    \caption{Comparison between close-set scene understanding framework and open-vocabulary scene understanding framework.}
    % \vspace{-0.6cm}
    \label{fig:framework_comp}
\end{figure*}

\begin{figure*}[htbp]
    % \vspace{-0.3cm}
    \begin{center}
    \includegraphics[width=1\linewidth]{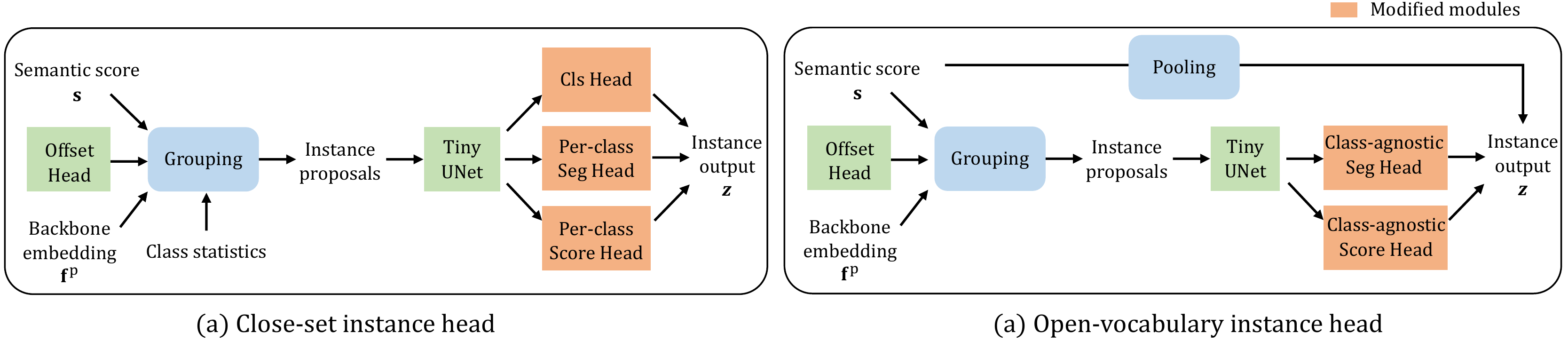}
    \end{center}
    % \vspace{-0.5cm}
    % captionsetup{font={small}}
    \caption{Comparison between close-set instance head and open-vocabulary instance head.}
    % \vspace{-0.6cm}
    \label{fig:instance_head}
\end{figure*}

\begin{table}[htbp]
% \vspace{-0.2cm}
    \centering
    \begin{small}
    \setlength\tabcolsep{8pt}
    \scalebox{1.0}{
        \begin{tabular}{c|c|c|c}
            \bottomrule[1pt]
            \multicolumn{3}{c|}{Components} & \multirow{2}{*}{mAP$_{50}$} \\
            \cline{1-3}
            \makecell{per-class seg head \\ and score head} & cls head & class statistics & \\
            \hline
            \checkmark & \checkmark & \checkmark & 61.8 \\
            & \checkmark & \checkmark & 62.0 \\
            & & \checkmark &  61.1 \\
            & & &  60.7 \\
            \toprule[0.8pt]
        \end{tabular}
    }
    \end{small}
    % \vspace{-0.3cm}
    \caption{Fully-supervised instance segmentation results of different SoftGroup variants upon ScanNet in terms of mAP$_{50}$.}
    \label{tab:softgroup}
    % \vspace{-0.3cm}
\end{table}

In this section, we elaborate on how to extend a close-set network to an open-vocabulary learner for semantic segmentation and instance segmentation. We employ sparse-convolution-based UNet~\cite{graham20183d} with a base hidden dimension of 16 as our backbone $\text{F}_{\text{3D}}$.

First, as illustrated in Fig.~\ref{fig:framework_comp} (a), the close-set network contains a learnable semantic head F$_\text{sem}$ that classifies a fixed number of categories. As discussed in Sec. {\color{red}3.2} in the main paper, to obtain an open-vocabulary model, we replace the semantic head F$_\text{sem}$ with a vision-language (VL) adapter F$_\theta$ and the category embedding $\mathbf{f}^l$ encoded by a fixed text encoder F$_\text{text}$. Note that the category embedding  $\mathbf{f}^l$ can be treated as replacing the weights of the classifier.
The category embedding $\mathbf{f}^l$ encodes semantic attributes of base classes in the training stage and encodes any desired categories during inference to achieve open-vocabulary semantic segmentation.

Further, as we follow SoftGroup~\cite{vu2022softgroup} to develop instance head F$_\text{ins}$, we modify the close-set designs in SoftGroup to obtain an open-vocabulary instance head. 
First, as shown in Fig.~\ref{fig:instance_head}, the seg head and the score head that produce per-class confidence in the vector form are modified to class-agnostic modules that produce a single scalar for each generated instance proposal. In this way, we can train these two heads without needing to know novel categories.
Second, the learnable cls head that predicts the classification scores of generated proposals is replaced by the proposal-level pooling of semantic scores $\mathbf{s}$, which can be extended to arbitrary categories.
Finally, the class statistics, such as the average number of points in an instance mask for each class, which assists proposal grouping, are removed to avoid leakage of novel class information. We empirically show that those modifications cause little  degradation of fully-supervised performance by 1.1\% mAP$_{50}$, as demonstrated in Table~\ref{tab:softgroup}. Note that we train the model from scratch rather than fine-tuning a supervised pretrained model, as SoftGroup does, to prevent leakage of novel classes during training. Additionally, we use a smaller hidden dimension size for the UNet backbone. Consequently, our reproduced performance differs from that in the original paper.

\subsection{Baseline Setups}
As mentioned in Sec. {\color{red}4.1} of the main paper, we follow LSeg~\cite{li2022languagedriven} to implement LSeg-3D as a baseline with UNet~\cite{graham20183d,choy20194d} backbone, vision-language adapter implemented by MLP and the CLIP~\cite{radford2021learning} ViT-B/16 text encoder. For the other two 3D zero-shot methods, 3DGenZ~\cite{michele2021generative} and 3DTZSL~\cite{cheraghian2020transductive}, we reproduce them with the same network and CLIP text embedding for fair comparisons. Specifically, for 3DGenZ~\cite{michele2021generative}, instead of training on samples that only contain base classes, we train it on the whole training dataset with points belonging to novel classes ignored during optimization. Besides, we remove calibrated stacking that aims to alleviate bias towards seen classes since it brings extremely minor performance gains in our implementations. As for 3DTZSL~\cite{cheraghian2020transductive} designed for object classification, we extend it to segmentation via learning with triplet loss on the point level instead of the sample level. We implement its projection net with 2 fully-connected layers and the Tanh activation function, the same as its paper claimed.

\subsection{Hyper-Parameter Configurations}

We train 19,216 iterations on ScanNet and 4,080 iterations on S3DIS for semantic segmentation. For instance segmentation, we train 24,020 iterations on ScanNet and 9,160 iterations on S3DIS. The learning rate is initialized as 0.004 with cosine decay. We adopt the AdamW~\cite{loshchilov2017decoupled} optimizer and run all experiments with 32 batch size on 8 NVIDIA V100 or NVIDIA A100.

For entity-level captions, we filter out some $\langle \mathbf{\hat{p}^e},\mathbf{t}^e \rangle$ pairs to guarantee the point set $\mathbf{\hat{p}^e}$ is small enough containing only a few entities. Specifically, we set the minimal points $\gamma$ as 100 and the ratio that controls the maximum number of points $\delta$ as 0.3.
As for the caption loss, we set $\alpha_1$, $\alpha_2$ and $\alpha_3$ as 0, 0.05 and 0.05 for scene-level $\mathcal{L}^{s}_\text{cap}$, view-level $\mathcal{L}^{v}_\text{cap}$ and entity-level loss $\mathcal{L}^{e}_\text{cap}$ for ScanNet, respectively. For S3DIS, we set $\alpha_1$, $\alpha_2$, and $\alpha_3$ as 0, 0.08, and 0.02 separately.

\subsection{Usage of Images}
For ScanNet, we use a 25,000-frame subset\footnotemark[4] from ScanNet images for captioning. For S3DIS, as each scene contains a widely varying number of images, we subsample its images to caption at most 50 images per scene. It is worth noting that some S3DIS scenes lack corresponding images; we consequently cannot provide language supervision for those scenes without images during training.

\footnotetext[4]{\url{https://kaldir.vc.in.tum.de/scannet_benchmark/documentation}}

\section{Analysis of PointCLIP for Scene Understanding}\label{sec:analysis}
In recent years, 2D open-vocabulary understanding~\cite{gu2021open,Hanoona2022Bridging,xu2021simple,li2022languagedriven} achieves unprecedented success driven by transferable vision-language models such as CLIP~\cite{radford2021learning} trained on large-scale image-caption pairs. Inspired by that success, PointCLIP~\cite{huang2022clip2point} has made the first attempt to transfer the knowledge of CLIP into the 3D domain for zero-shot and few-shot object classification tasks. PointCLIP projects 3D point clouds into 2D multi-view depth maps and leverages CLIP to process multi-view depth images to obtain  predictions. Finally, the predictions are assembled into 3D predictions. Though some progress has been made in object-level understanding, our experimental results show that PointCLIP is not suitable for scene-level understanding tasks with poor performance and heavy inference overheads.
%we empirically show this stream of works is inapplicable for scene understanding tasks due to poor performance and heavy inference overheads. 

\noindent\textbf{Task-specific modifications.} To extend PointCLIP for 3D scene understanding, we make the following modifications. 
First, we follow the state-of-the-art 2D open-vocabulary semantic segmentation method MaskCLIP~\cite{zhou2022maskclip} to modify the attentive pooling layer of CLIP's vision encoder for obtaining pixel-wise dense predictions. Second, instead of using self-rendered images, we utilize collected depth images captured by depth sensors since they are realistic with more accurate depth values.
We also explore utilizing collected RGB images to avoid modal gaps caused by using depth images.
Finally, to assemble multi-view 2D results into 3D, other than voting to get object-wise predictions, we back-project all multi-view image predictions into 3D space via 3D geometry and assign predictions to each point of 3D scenes by searching nearest neighbors in back-projected 3D point clouds.

% To fit the one-dimension depth image to three-dimension feature map for CLIP processing, it simply repeat the depth values for all three channels.
% Besides, it hand-craft a text prompt for encoding category names

\noindent\textbf{Results.}
\begin{table}[htbp]
% \vspace{-0.2cm}
    \centering
    \begin{small}
    \setlength\tabcolsep{8pt}
    \scalebox{1.0}{
        \begin{tabular}{c|c|c|c}
            \bottomrule[1pt]
            Input & 2D mIoU & 3D mIoU & latency (ms) \\
            \hline
            depth images & 02.2 & 01.7 & 1667\\
            RGB images & 17.8 & 17.2 & 1667 \\
            \toprule[0.8pt]
        \end{tabular}
    }
    \end{small}
    % \vspace{-0.3cm}
    \caption{Results of zero-shot 3D semantic segmentation using PointCLIP on ScanNet.}
    \label{tab:pointclip}
    % \vspace{-0.3cm}
\end{table}
As shown in Table~\ref{tab:pointclip}, with depth images as input, the modified PointCLIP obtains only 2.2\% mIoU on 2D semantic segmentation with 5,436 validation samples of ScanNet. The assembled 3D prediction only attains 1.7\% mIoU on 312 samples, which is very close to random guesses. When alternated to use RGB images as input, the performance lifts to 17.8\% mIoU on 2D and 17.2\% mIoU on 3D, demonstrating that using RGB images can avoid annoying modal gaps. However, the performance is still moderate, which suggests this projection-based stream of work is sub-optimal for tackling 3D scene understanding tasks. 
Though further fine-tuning on seen categories might benefit model performance, this line of research has a key limitation: by projecting 3D data to 2D, it suffers from information loss and makes the model unable to directly learn from information-rich 3D data.

In addition, to assess the model efficiency, we use latency to measure the execution speed of model inference on a single GeForce RTX 2080Ti. As shown in Table~\ref{tab:pointclip}, PointCLIP takes an average of 1667ms to process images of one 3D scene, which is rather costly, not to mention the post-processing time for back-projection and results ensemble.
Instead, our 3D network only costs 83ms to process one 3D sample, which is 20 times more efficient than PointCLIP.

In sum, the poor zero-shot performance, information loss from projection, and heavy computation costs render this line of work not suitable for 3D scene understanding and prevent us from exploring further on this stream of work.

\section{Additional Experimental Results}\label{sec:result}

\subsection{Re-partition Experiments}

\begin{table}[htbp]
% \vspace{-0.2cm}
    \centering
    \begin{small}
    \setlength\tabcolsep{8pt}
    \scalebox{1.0}{
        \begin{tabular}{l|c|c}
            \bottomrule[1pt]
            \multirow{2}{*}{Splits}  & \multicolumn{2}{c}{hIoU / mIoU$^\mathcal{B}$ / mIoU$^\mathcal{N}$} \\
            \cline{2-3}
            & LSeg-3D~\cite{li2022languagedriven} & Ours \\
            \hline
            random-sample 1 & 00.0 / 61.7 / 00.0 & \textbf{65.3} / \textbf{68.3} / \textbf{62.4} \\
            random-sample 2 & 00.0 / 48.5 / 00.0 & \textbf{53.1} / \textbf{70.1} / \textbf{42.7} \\
            random-sample 3 & 00.3 / 66.1 / 00.2 & \textbf{60.9} / \textbf{69.2} / \textbf{54.5} \\
            frequency-sample & 00.0 / 68.7 / 00.0 & \textbf{62.6} / \textbf{69.0} / \textbf{57.3} \\
            \toprule[0.8pt]
        \end{tabular}
    }
    \end{small}
    % \vspace{-0.3cm}
    \caption{Results of re-sampled base and novel categories.}
    \label{tab:partitions}
    % \vspace{-0.3cm}
\end{table}

To ensure the reliability of results, we randomly re-sample base and novel categories three times and sample it based on class frequency for the B15/N4 ScanNet semantic segmentation task. As shown in Table~\ref{tab:partitions}, our method consistently exceeds LSeg-3D baseline among four different splits by a large margin of $53.1\%\sim 65.3\%$ hIoU, which reveals the robustness of our methods in handling different novel classes.

\subsection{Per-class Results}

\begin{table*}[htbp]
    \centering
    % \vspace{-0.3cm}
    \begin{small}
      \scalebox{1.0}{
        \setlength{\tabcolsep}{1.3mm}{
        \begin{tabular}{c|c|ccccccccccccccccccc}
            \bottomrule[1pt]
            Task & Partition & \rotatebox[origin=c]{90}{wall} & \rotatebox[origin=c]{90}{floor} & \rotatebox[origin=c]{90}{cabinet} & \rotatebox[origin=c]{90}{bed} & \rotatebox[origin=c]{90}{chair} & \rotatebox[origin=c]{90}{sofa} & \rotatebox[origin=c]{90}{table} & \rotatebox[origin=c]{90}{door} & \rotatebox[origin=c]{90}{window} & \rotatebox[origin=c]{90}{ bookshelf } & \rotatebox[origin=c]{90}{picture} & \rotatebox[origin=c]{90}{counter} & \rotatebox[origin=c]{90}{desk} & \rotatebox[origin=c]{90}{curtain} & \rotatebox[origin=c]{90}{fridge} & \rotatebox[origin=c]{90}{shower c.} & \rotatebox[origin=c]{90}{toilet} & \rotatebox[origin=c]{90}{sink} & \rotatebox[origin=c]{90}{bathtub} \\
            \hline
            \multirow{3}{*}{Sem.} & B15/N4 & 84.6 & 95.0 & 64.9 & 81.1 & 87.9 & {\cellcolor{myblue2}75.9} & 72.2 & 61.9 & 62.1 & {\cellcolor{myblue2}69.5} & 30.9 & 60.1 & {\cellcolor{myblue2}46.5} & 70.7 & 50.5 & 66.1 & {\cellcolor{myblue2}56.8} & 59.0 & 81.7 \\
            & B12/N7 & 84.7 & 95.1 & 65.3 & {\cellcolor{myblue2}57.8} & {\cellcolor{myblue2}44.2} & 75.9 & {\cellcolor{myblue2}34.5} & 62.5 & 62.3 & {\cellcolor{myblue2}62.1} & {\cellcolor{myblue2}20.5} & 57.8 & 61.4 & 72.4 & 47.9 & 64.9 & 85.9 & {\cellcolor{myblue2}28.4} & {\cellcolor{myblue2}69.6} \\
            & B10/N9 & 83.8 & 95.2 & 64.3 & 80.9 & 88.0 & 78.5 & 73.2 & 60.6 & 61.5 & {\cellcolor{myblue2}68.6} & {\cellcolor{myblue2}17.7} & {\cellcolor{myblue2}23.4} & {\cellcolor{myblue2}51.3} & 70.6 & {\cellcolor{myblue2}25.7} & {\cellcolor{myblue2}38.2} & {\cellcolor{myblue2}51.3} & {\cellcolor{myblue2}27.3} & {\cellcolor{myblue2}61.7} \\
            \hline
            \multirow{3}{*}{Inst.} & B13/N4 & $-$ & $-$ & 50.5 & 77.0 & 82.9 & {\cellcolor{myblue2}43.4} & 75.4 & 49.0 & 46.0 & {\cellcolor{myblue2}43.7} & 46.5 & 33.7 & {\cellcolor{myblue2}23.2} & 54.1 & 49.6 & 56.0 & {\cellcolor{myblue2}97.8} & 47.5 & 85.8 \\
            & B10/N7 & $-$ & $-$ & 53.7 & {\cellcolor{myblue2}62.7} & {\cellcolor{myblue2}11.2} & 70.5 & {\cellcolor{myblue2}27.2} & 47.7 & 45.7 & {\cellcolor{myblue2}30.0} & {\cellcolor{myblue2}01.5} & 39.9 & 40.8 & 50.6 & 68.6 & 84.6 & 92.9 & {\cellcolor{myblue2}24.6} & {\cellcolor{myblue2}00.0} \\
            & B8/N9 & $-$ & $-$ & 45.1 & 77.4 & 82.2 & 84.2 & 74.2 & 48.9 & 51.0 & {\cellcolor{myblue2}30.0} & {\cellcolor{myblue2}00.5} & {\cellcolor{myblue2}02.1} & {\cellcolor{myblue2}16.8} & 44.9 & {\cellcolor{myblue2}28.3} & {\cellcolor{myblue2}35.1} & {\cellcolor{myblue2}94.3} & {\cellcolor{myblue2}16.6} & {\cellcolor{myblue2}00.0} \\
            \toprule[1pt]
        \end{tabular}}}
    \end{small}
    \caption{Per-class results of 3D open-vocabulary scene understanding on ScanNet. Performance on novel class are marked in \colorbox{myblue2}{blue}.}
    \label{tab:per_class_scannet}
\end{table*}

\begin{table*}[htbp]
    \centering
    % \vspace{-0.3cm}
    \begin{small}
      \scalebox{1.0}{
        \setlength{\tabcolsep}{2mm}{
        \begin{tabular}{c|c|cccccccccccc}
            \bottomrule[1pt]
            Task & Partition & \rotatebox[origin=c]{90}{ ceiling } & \rotatebox[origin=c]{90}{floor} & \rotatebox[origin=c]{90}{wall} & \rotatebox[origin=c]{90}{beam} & \rotatebox[origin=c]{90}{column} & \rotatebox[origin=c]{90}{window} & \rotatebox[origin=c]{90}{door} & \rotatebox[origin=c]{90}{table} & \rotatebox[origin=c]{90}{chair} & \rotatebox[origin=c]{90}{sofa} & \rotatebox[origin=c]{90}{ bookcase } & \rotatebox[origin=c]{90}{board} \\
            \hline
            \multirow{2}{*}{Sem.} & B8/N4 & 93.9 & 97.8 & 82.9 & 00.0 & 17.2 & {\cellcolor{myblue2}15.6} & 53.7 & {\cellcolor{myblue2}35.8} & 86.3 & {\cellcolor{myblue2}05.3} & {\cellcolor{myblue2}37.3} & 43.3 \\
            & B6/N6 & 93.7 & {\cellcolor{myblue2}79.1} & 80.1 & 00.1 & 28.5 & {\cellcolor{myblue2}24.1} & {\cellcolor{myblue2}08.4} & {\cellcolor{myblue2}37.6} & 87.0 & 54.0 & {\cellcolor{myblue2}24.0} & {\cellcolor{myblue2}06.9} \\
            \hline
            \multirow{2}{*}{Inst.} & B8/N4 & 89.5 & 100.0 & 50.8 & 00.0 & 35.3 & {\cellcolor{myblue2}36.2} & 60.5 & {\cellcolor{myblue2}00.1} & 84.6 & {\cellcolor{myblue2}01.9} & {\cellcolor{myblue2}00.8} & 59.4 \\
            & B6/N6 & 89.5 & {\cellcolor{myblue2}60.2} & 17.9 & 00.0 & 41.5 & {\cellcolor{myblue2}10.2} & {\cellcolor{myblue2}02.1} & {\cellcolor{myblue2}00.6} & 86.2 & 45.1 & {\cellcolor{myblue2}00.1} & {\cellcolor{myblue2}02.2} \\
            \toprule[1pt]
        \end{tabular}}}
    \end{small}
    \caption{Per-class results of 3D open-vocabulary scene understanding on S3DIS.}
    \label{tab:per_class_s3dis}
\end{table*}

We present per-category performances of our open-vocabulary 3D scene understanding framework on semantic and instance segmentation.
As shown in Table~\ref{tab:per_class_scannet} and Table~\ref{tab:per_class_s3dis}, novel classes generally perform worse than base classes without annotation supervision. With the space of novel categories enlarged (\eg from B15/N4 to B12/N7 partition), the performance on novel classes degrades (\eg `bookshelf' obtains 7.4\% mIoU drop from B15/N4 to B12/N7 partition on semantic segmentation) due to the insufficient seen-category data to tune the model.
%due to more confusion among different novel classes.

\subsection{Error Bar}
\begin{table*}[htbp]
    % \vspace{-0.1cm}
    \centering
    \begin{small}
    \setlength\tabcolsep{2pt}
    \scalebox{1.0}{
        \begin{tabular}{c|ccc|ccc|ccc|ccc|ccc}
            \bottomrule[1pt]
            \multirow{3}{*}{Round}  & \multicolumn{9}{c|}{ScanNet} & \multicolumn{6}{c}{S3DIS} \\
            \cline{2-16}
            & \multicolumn{3}{c|}{B15/N4} & \multicolumn{3}{c|}{B12/N7} & \multicolumn{3}{c|}{B10/N9} & \multicolumn{3}{c|}{B8/N4} & \multicolumn{3}{c}{B6/N6} \\
            \cline{2-16}
            & hIoU & mIoU$^\mathcal{B}$ & mIoU$^\mathcal{N}$ & hIoU & mIoU$^\mathcal{B}$ & mIoU$^\mathcal{N}$ & hIoU & mIoU$^\mathcal{B}$ & mIoU$^\mathcal{N}$ & hIoU & mIoU$^\mathcal{B}$ & mIoU$^\mathcal{N}$ & hIoU & mIoU$^\mathcal{B}$ & mIoU$^\mathcal{N}$ \\
            \hline
            1 & 66.3 & 68.4 & 64.2 & 54.3 & 69.5 & 44.6 & 52.8 & 76.2 & 40.6 & 33.2 & 58.2 & 23.3 & 39.4 & 57.2 & 30.0 \\
            2 & 65.2 & 68.6 & 62.2 & 54.8 & 69.7 & 45.2 & 53.3 & 75.6 & 40.9 & 37.0 & 59.5 & 26.9 & 39.5 & 55.1 & 30.8 \\
            3 & 64.5 & 67.8 & 60.8 & 59.7 & 69.2 & 48.0 & 53.2 & 76.6 & 40.8 & 33.7 & 59.4 & 23.5 & 36.5 & 54.3 & 27.5 \\
            \hline
            Average & 65.3 & 68.3 & 62.4 & 55.3 & 69.5 & 45.9 & 53.1 & 76.2 & 40.8 & 34.6 & 59.0 & 24.5 & 38.5 & 55.5 & 29.4 \\
            Std & 00.9 & 00.4 & 01.7 & 01.3 & 00.2 & 01.8 & 00.3 & 00.5 & 00.2 & 02.1 & 00.7 & 02.0 & 01.7 & 01.5 & 01.7 \\
            \toprule[0.8pt]
        \end{tabular}
     }
    \end{small}
    % \vspace{-0.4cm}
    \caption{Repeat results for open-vocabulary 3D semantic segmentation on ScanNet and S3DIS in terms of hIoU, mIoU$^\mathcal{B}$ and mIoU$^\mathcal{N}$.}
    \label{tab:error_bar_sem}
    % \vspace{-0.2cm}
\end{table*}

\begin{table*}[htbp]
    % \vspace{-0.1cm}
    \centering
    \begin{small}
    \setlength\tabcolsep{2pt}
    \scalebox{1.0}{
        \begin{tabular}{c|ccc|ccc|ccc|ccc|ccc}
            \bottomrule[1pt]
            \multirow{3}{*}{Round}  & \multicolumn{9}{c|}{ScanNet} & \multicolumn{6}{c}{S3DIS} \\
            \cline{2-16}
            & \multicolumn{3}{c|}{B13/N4} & \multicolumn{3}{c|}{B10/N7} & \multicolumn{3}{c|}{B8/N9} & \multicolumn{3}{c|}{B8/N4} & \multicolumn{3}{c}{B6/N6} \\
            \cline{2-16}
            & hAP$_{50}$ & mAP$_{50}^\mathcal{B}$ & mAP$_{50}^\mathcal{N}$ & hAP$_{50}$ & mAP$_{50}^\mathcal{B}$ & mAP$_{50}^\mathcal{N}$ & hAP$_{50}$ & mAP$_{50}^\mathcal{B}$ & mAP$_{50}^\mathcal{N}$ & hAP$_{50}$ & mAP$_{50}^\mathcal{B}$ & mAP$_{50}^\mathcal{N}$ & hAP$_{50}$ & mAP$_{50}^\mathcal{B}$ & mAP$_{50}^\mathcal{N}$ \\
            \hline
            1 & 54.9 & 58.1 & 52.0 & 33.1 & 52.5 & 24.1 & 34.5 & 62.1 & 23.9 & 19.3 & 59.2 & 11.5 & 10.9 & 49.2 & 06.1 \\
            2 & 56.7 & 57.9 & 55.5 & 28.4 & 55.1 & 19.1 & 37.5 & 63.8 & 26.5 & 9.2 & 57.4 & 05.0 & 19.8 & 46.7 & 12.6 \\
            3 & 55.0 & 59.5 & 51.1 & 32.1 & 56.3 & 22.5 & 35.7 & 63.5 & 24.8 & 16.8 & 60.0 & 09.7 & 17.4 & 44.9 & 10.8 \\
            \hline
            Average & 55.5 & 58.5 & 52.9 & 31.2 & 54.6 & 21.9 & 35.9 & 63.1 & 25.1 & 15.0 & 59.0 & 08.6 & 16.0 & 46.9 & 09.8 \\
            Std & 01.0 & 00.9 & 02.3 & 02.5 & 01.9 & 02.6 & 01.5 & 00.9 & 01.3 & 04.3 & 01.1 & 02.7 & 04.6 & 02.2 & 03.4 \\
            \toprule[0.8pt]
        \end{tabular}
     }
    \end{small}
    % \vspace{-0.4cm}
    \caption{Repeat results for open-vocabulary 3D instance segmentation on ScanNet and S3DIS in terms of hAP$_{50}$, mAP$_{50}^\mathcal{B}$ and mAP$_{50}^\mathcal{N}$.}
    \label{tab:error_bar_inst}
    % \vspace{-0.2cm}
\end{table*}

Here, to show the robustness of our experimental results, we repeat the experiments on open-vocabulary semantic and instance segmentation three times and report their average along with standard deviation. As shown in Table~\ref{tab:error_bar_sem} and Table~\ref{tab:error_bar_inst}, the results on base classes are slightly more stable than novel classes with lower standard deviations, which demonstrates the higher confidence uncertainty of novel class predictions. Besides, results on ScanNet are more stable than S3DIS, which indicates that the sample size and diversity contribute a lot to the performance stability.

\subsection{Equipping Fully-Supervised Model with Point-Caption Supervision.}

As demonstrated in Table~\ref{tab:fully_sup}, fully-supervised models equipped with caption supervision loss perform similarly to those without it, as they already have access to annotations for all categories. In this scenario, our language supervision neither hinders nor enhances fully-supervised performance, validating our fairness in using the fully-supervised model for comparison in the main paper.

\begin{table}[htbp]
\vspace{-0.4cm}
    \centering
    \begin{small}
    \setlength\tabcolsep{3pt}
    \scalebox{1.0}{
        \begin{tabular}{c|c|c|c|c}
            \bottomrule[1pt]
            \multirow{2}{*}{Method} & \multirow{2}{*}{mIoU} & \multicolumn{3}{c}{mIoU$^\mathcal{B}$ / mIoU$^\mathcal{N}$}\\
            \cline{3-5}
            & & B15/N4 & B12/N7 & B10/N9 \\
            \hline
            Fully-Sup. & 70.62 & 68.4 / 79.1 & 70.0 / 71.8 & 75.8 / 64.9 \\
            Fully-Sup. + Caption & 70.82 & 68.7 / 78.9 & 70.3 / 71.7 & 76.7 / 64.6\\
            \toprule[0.8pt]
        \end{tabular}
    }
    \end{small}
    \caption{Fully-supervised results equipped with point-caption supervision.}
    \label{tab:fully_sup}
\end{table}

\subsection{Combination of Caption Supervisions.} The combination of three captions, including the scene-level caption, can result in a 0.6\% increase in hIoU, as shown in Table~\ref{tab:loss_weight}. However, finding such a right balance between these captions requires sophisticated loss trade-off techniques that are not universally applicable across different datasets. Therefore, the scene-level caption is not used in our paper for the sake of generalization. Further studies on effectively combining caption supervisions would be a future investigation.

\begin{table}[htbp]
\vspace{-0.4cm}
    \centering
    \begin{small}
    \setlength\tabcolsep{4pt}
    \scalebox{1.0}{
        \begin{tabular}{ccc|c}
            \bottomrule[1pt]
            $\alpha_1$(scene) & $\alpha_2$(view) & $\alpha_3$(entity) & hIoU / mIoU$^\mathcal{B}$ / mIoU$^\mathcal{N}$ \\
            \hline
            0.000 & 0.050 & 0.050 & 65.3 / 68.3 / 62.4\\
            0.033 & 0.033 & 0.033 & 64.6 / \textbf{69.0} / 60.8 \\
            0.010 & 0.045 & 0.045 & \textbf{65.9} / 68.2 / \textbf{63.8} \\
            \toprule[0.8pt]
        \end{tabular}
    }
    \end{small}
    \caption{Ablation for caption loss weights on ScanNet B15/N4.}
    \label{tab:loss_weight}
\end{table}

\section{Caption Examples}\label{sec:example}

\begin{figure*}[htbp]
    % \vspace{-0.3cm}
    \begin{center}
    \includegraphics[width=1\linewidth]{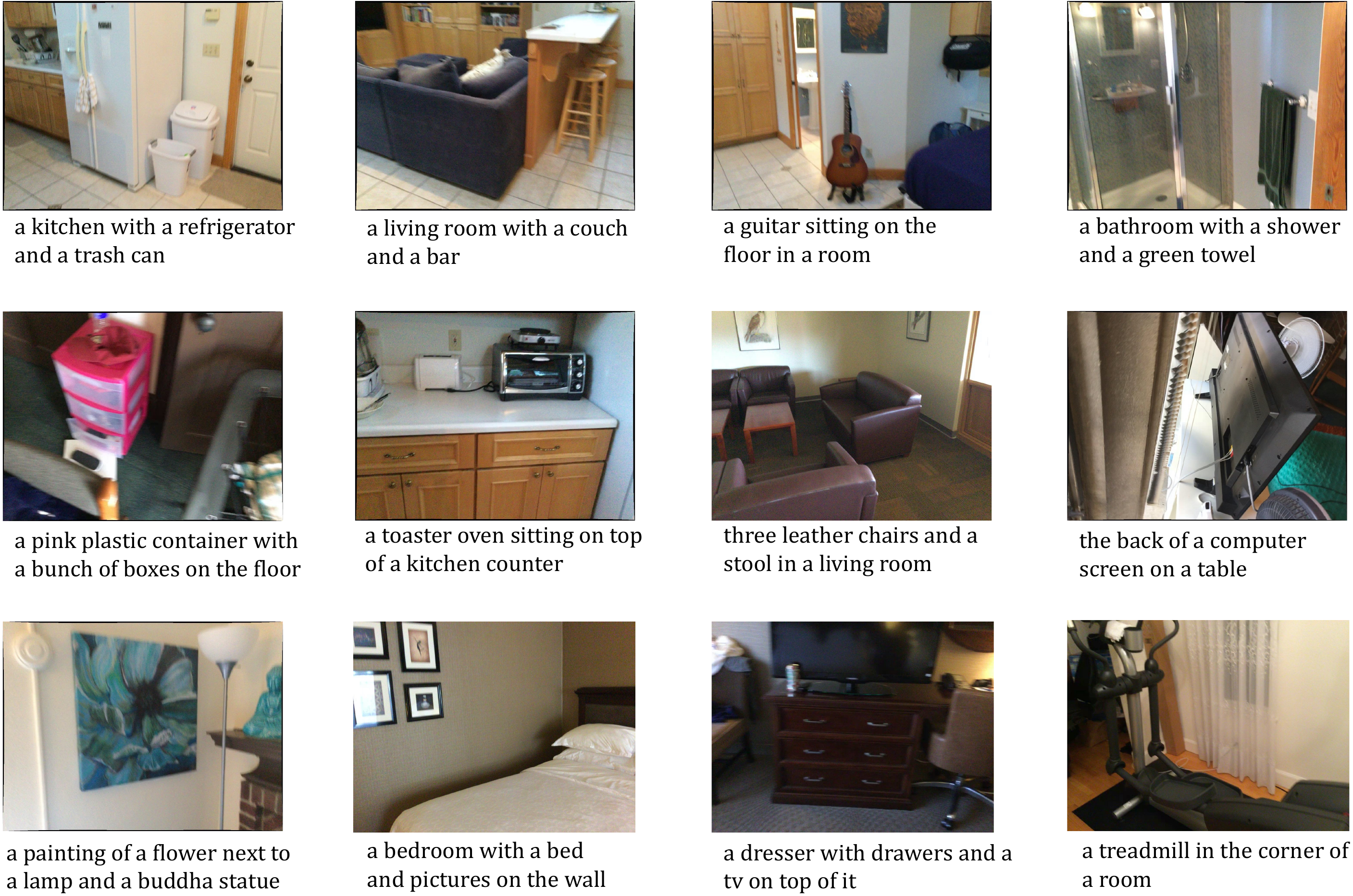}
    \end{center}
    \caption{Examples of image-caption pairs by image-captioning model ViT-GPT2~\cite{vit-gpt2}.}
    \label{fig:image-caption}
\end{figure*}

\begin{figure*}[htbp]
    \begin{center}
    \includegraphics[width=1\linewidth]{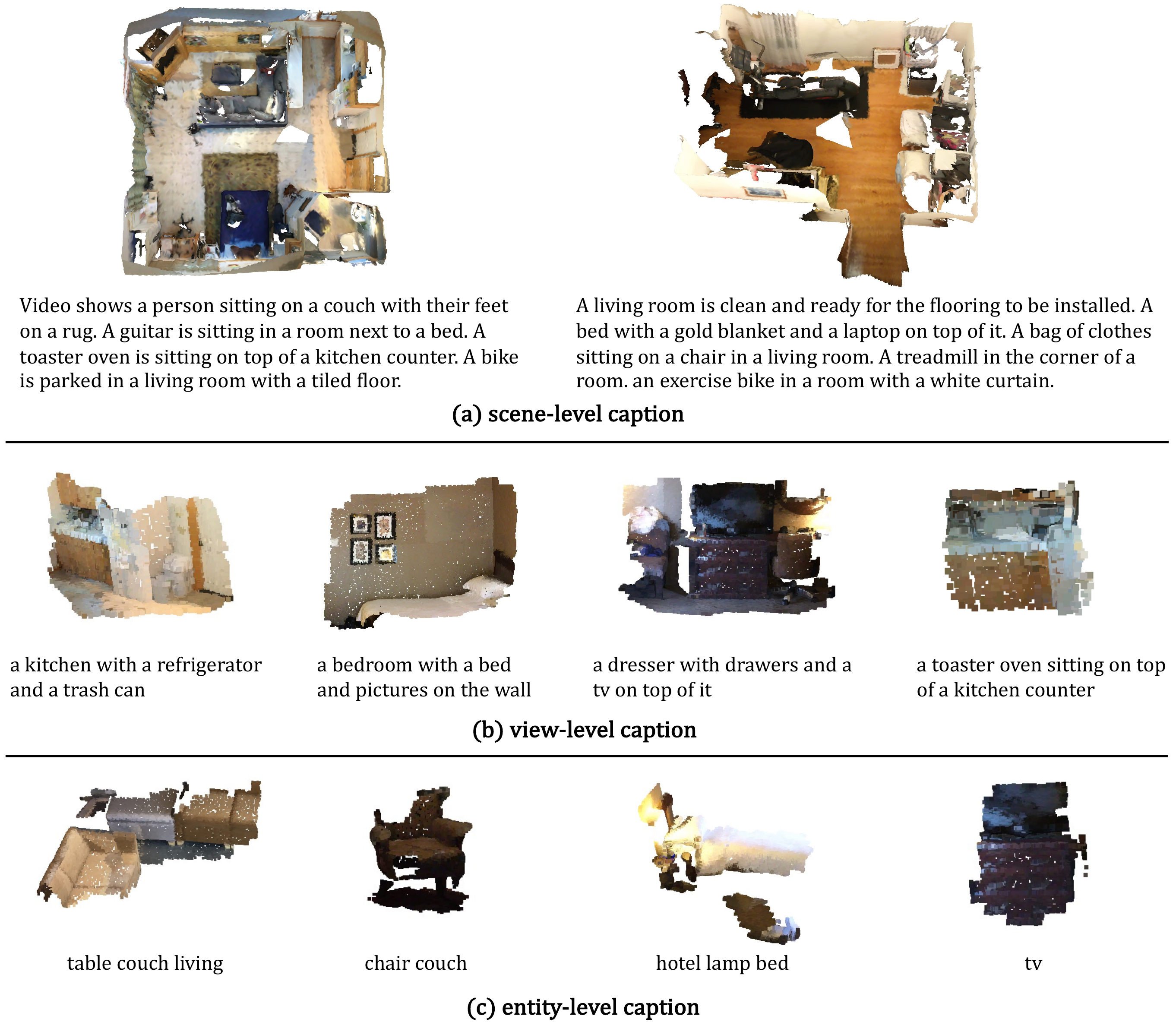}
    \end{center}
    \caption{Examples of hierarchical point-caption pairs from ScanNet~\cite{dai2017scannet}}.
    \label{fig:point-caption}
\end{figure*}

In this section, we present examples of image-caption pairs obtained by vision-language (VL) foundation models and examples of hierarchical associated point-caption pairs.
As illustrated in Fig.~\ref{fig:image-caption}, image captions describe main entities of images along with room types (\eg kitchen), texture (\eg leather), color (\eg green) or spatial relationships (\eg on top of), conveying rich semantic clues with large vocabulary size. Moreover, uncommon classes such as  `buddha statue' are also correctly detected, reflecting the generalizability of existing VL foundation models and semantic comprehensiveness of generated captions.

With obtained image-caption pairs, we are capable to associate 3D points and captions hierarchically leveraging geometric constraints between 3D point clouds and multi-view images. As shown in Fig.~\ref{fig:point-caption} (a), the scene-level caption describes each area/room (\eg kitchen, living room) in the whole scene with abundant vocabulary, providing semantic-rich language supervision. View-level caption in Fig.~\ref{fig:point-caption} (b) focuses on single view frustums of the 3D point cloud, capturing more local details with elaborate text descriptions, which enables the model to learn region-wise vision-semantic relationships. Additionally, as shown in Fig.~\ref{fig:point-caption} (c), the entity-level caption covers only a few entities in small 3D point sets with concrete words, providing more fine-grained supervisions to learn object-level understanding and localization.

\section{Qualitative Results}\label{sec:vis}

\begin{figure*}[htbp]
    % \vspace{-0.3cm}
    \begin{center}
    \includegraphics[width=1\linewidth]{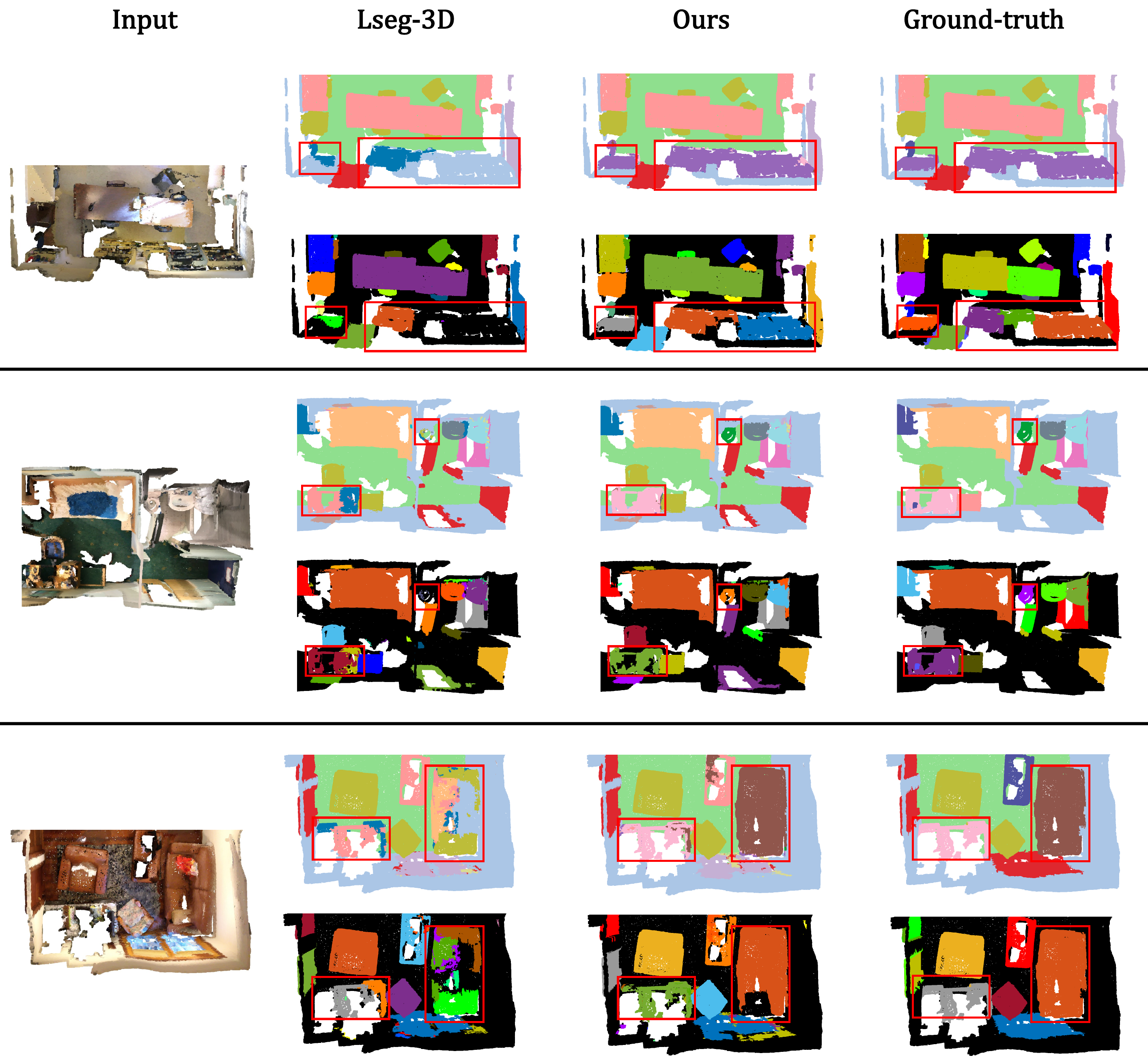}
    \end{center}
    % \vspace{-0.5cm}
    % captionsetup{font={small}}
    \caption{Qualitative results of open-vocabulary semantic segmentation and instance segmentation. In each example, the first row illustrates the semantic masks and the second row shows the instance masks. Novel classes are highlighted in red bounding boxes.}
    % \vspace{-0.6cm}
    \label{fig:vis}
\end{figure*}

Here, we provide some qualitative results on open-vocabulary semantic segmentation and instance segmentation as illustrated in Fig.~\ref{fig:vis}. Compared to the LSeg-3D baseline that always confuses unseen classes as seen classes, our framework successfully recognizes novel categories with accurate semantic masks, which shows our point-caption association injects rich semantic concepts into the 3D network. Additionally, the instance prediction masks of our framework are also accurate, while the LSeg-3D baseline misses novel objects or predicts incomplete object masks. It reflects the strong generalized localization ability of our framework.

\section{Limitation and Open Problems}\label{sec:limitation}
Although our language-driven open-vocabulary 3D scene understanding framework introduces rich semantic concepts for learning adequate visual-semantic relationships, it still suffers from limitations in the following aspects.
First comes the calibration problem that the model tends to produce over-confident predictions on base classes, which lies in both semantic and instance segmentation tasks.
For semantic segmentation, though the binary head is developed to calibrate semantic scores for in-domain open-vocabulary scene understanding, it fails to rectify predictions for out-of-domain transfer tasks. Trained on the dataset-specific base/novel partition, the binary head is hard to generalize to other datasets with data distribution shifts, which encourages us to design more transferable score calibration modules in the future. As for the instance segmentation task, though we largely address the localization problem for novel classes through fine-grained point-caption pairs, the calibration problem also exists in the proposal grouping process, where objects of novel classes cannot group well and probably obtain incomplete instance masks. We also leave it as a challenge that needs to be resolved further.

The second problem is that S3DIS achieves slightly worse open-vocabulary performance than ScanNet, largely due to its limited sample size and diversity, as well as much fewer point-caption associations. Inspired by our zero-shot transfer results, we believe it is an appealing alternative to pre-train on a large dataset with rich semantic information and then fine-tune it on the small-scale dataset, which we leave for future study.

\end{document}